\documentclass[11pt,letterpaper]{article}
\usepackage[margin=1in]{geometry}
\usepackage[T1]{fontenc}
\usepackage{lmodern}

\usepackage{amsmath,amsfonts,bm}

\def\secref#1{section~\ref{#1}}

\def\eqref#1{equation~\ref{#1}}

\def\1{\bm{1}}

\DeclareMathAlphabet{\mathsfit}{\encodingdefault}{\sfdefault}{m}{sl}
\SetMathAlphabet{\mathsfit}{bold}{\encodingdefault}{\sfdefault}{bx}{n}

\usepackage[round,authoryear]{natbib}
\usepackage[hyphens]{url}
\usepackage{graphicx}
\usepackage[table]{xcolor}
\usepackage{booktabs,tabularx,multirow}
\usepackage{pifont,float,placeins,caption}
\usepackage{microtype}
\usepackage{hyperref}
\hypersetup{colorlinks=true,linkcolor=blue,citecolor=blue,urlcolor=blue,
  pdftitle={DA-WAM: Decision-Aligned Future Latents for Driving World Models},
  pdfauthor={Ruiguo Zhong, Xiaolong Chen, Benshan Ma, Lang Zhang, Mingyue Feng, Yaonong Wang, Pei Liu, Jun Ma}}
\newcommand{\method}{DA-WAM}

\newcommand{\methodfigure}[3]{\includegraphics[width=#2]{#1}}
\providecommand{\cmark}{\ding{51}}
\providecommand{\xmark}{\ding{55}}
\newcommand{\highlightrow}{\rowcolor[HTML]{E8F2FA}}

\definecolor{ablationBlue}{RGB}{232,240,248}
\definecolor{ablationGray}{RGB}{85,95,108}
\newlength{\ablationLabelWidth}
\newlength{\mainTableWidth}
\title{DA-WAM: Decision-Aligned Future Latents for Driving World Models}
\author{%
  \normalsize Ruiguo Zhong$^1$, Xiaolong Chen$^1$, Benshan Ma$^1$, Lang Zhang$^2$\thanks{Project leader.}\\
  \normalsize Mingyue Feng$^2$, Yaonong Wang$^2$, Pei Liu$^1$\thanks{Corresponding author.}, Jun Ma$^{1,3}$\\[0.5em]
  \small $^1$The Hong Kong University of Science and Technology (Guangzhou)\\
  \small $^2$Leapmotor\\
  \small $^3$The Hong Kong University of Science and Technology\\[0.3em]
  \small \texttt{rzhong151@connect.hkust-gz.edu.cn}\\[0.3em]
  \small Project: \url{https://github.com/LeapWM/da-wam}
}
\date{}
\begin{document}

\maketitle

\begin{abstract}

World models for autonomous driving should anticipate the consequences that matter for driving decisions rather than reconstruct every detail of future scenes. However, predictive representations alone are not guaranteed to distinguish candidate actions that lead to different planning outcomes.
We propose \method{}, a decision-aligned world action model that couples a Joint-Embedding Predictive Architecture (JEPA) with candidate-level trajectory scoring. During joint training, the JEPA objective grounds future representations in observed scene dynamics, while the planning objectives encourage these representations to capture distinctions relevant to safety and utility. An action-conditioned predictor generates a distinct future latent for each candidate trajectory, and a multi-head trajectory scorer uses the corresponding latent to estimate planning factors and overall utility. In addition, safety-critical post-training with hard negatives further improves discrimination between geometrically similar trajectories with different safety outcomes.
\method{} achieves state-of-the-art performance among learned camera-based planners on NAVSIM-v1, NAVSIM-v2, and Bench2Drive. Ablation studies demonstrate the effectiveness of each component in driving decision-making.

\end{abstract}

\section{Introduction}

Safe and effective autonomous driving requires reasoning about how the scene would evolve under each candidate ego action. World models address this challenge by predicting future visual or latent states from current observations and proposed actions~\cite{wang2024driving,li2025enhancing,zheng2025world4drive}. Yet the fundamental question is not simply whether a model can predict the future, but whether its predictions are decision-informative: do the predicted consequences of each candidate action directly determine how that trajectory is evaluated? To realize the full potential of world models, future prediction must therefore be tightly coupled with decision making, such that each candidate is evaluated against the future predicted specifically for that action.

\begin{figure}[t]
    \centering
    \includegraphics[width=0.96\linewidth]{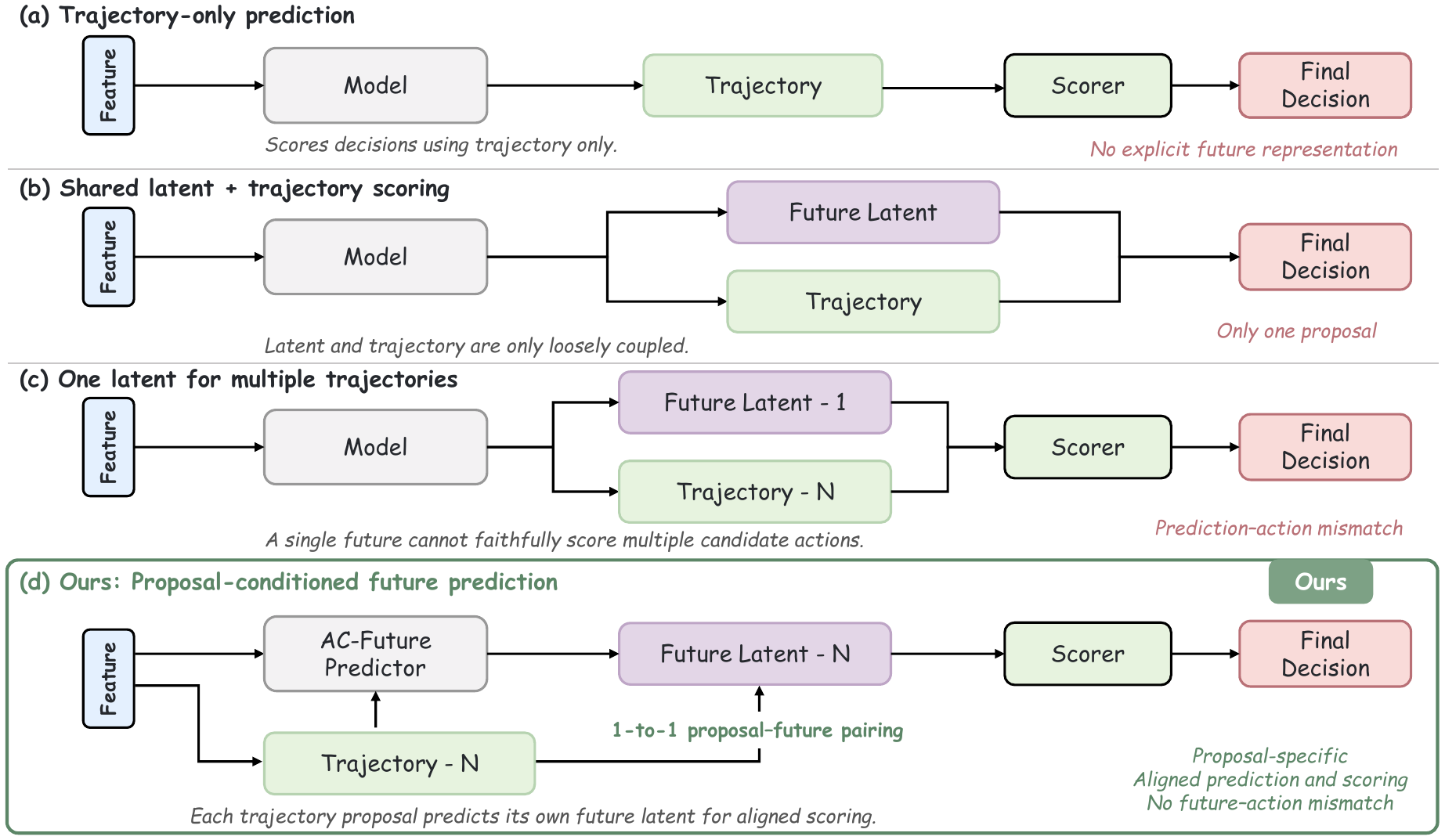}
    \caption{\textbf{Candidate--representation coupling in driving decisions.} Schematic comparison of (a) trajectory prediction without an explicit future-representation input to the scorer, (b) joint trajectory and latent prediction without explicit candidate-wise pairing, (c) multiple candidates evaluated using a shared future representation, and (d) \method{}, which computes a candidate-conditioned representation for each trajectory and scores the corresponding trajectory--representation pair. The scorer receives the trajectory together with its associated representation and scene context. The panels illustrate conditioning structures rather than an exhaustive classification of prior work.}
    \label{fig:pipeline_comparison}
\end{figure}

Existing approaches pursue future prediction along two broad directions, but neither direction alone guarantees this coupling. Methods such as DriveWorld~\cite{min2024driveworld}, LAW~\cite{li2025enhancing}, Drive-JEPA~\cite{wang2026drivejepa}, and Latent-WAM~\cite{wang2026latentwam} use temporal prediction primarily to strengthen scene representations or trajectory learning. Although these approaches learn rich predictive features, predictive pretraining and planner optimization are often stage-separated; fixed or frozen encoders consequently cannot adapt their future representations to the specific demands of trajectory scoring. A second line of work brings predicted futures more directly into planning. Drive-WM~\cite{wang2024driving} generates future views for planning, WoTE~\cite{li2025end} evaluates trajectories through a BEV world model, World4Drive~\cite{zheng2025world4drive} reasons over intention-conditioned latent futures, and DriveFuture~\cite{hong2026drivefuture} conditions its planner on a predicted future latent. These methods move closer to using future predictions at decision time, yet predicted states may still be shared, pooled, or only weakly associated with individual candidates. As summarized in Fig.~\ref{fig:pipeline_comparison}, these designs weaken the correspondence between an action and its predicted consequence. The scorer may therefore rely primarily on geometric cues rather than the scene-conditioned future content that distinguishes safe from unsafe outcomes.

We argue that the planning value of a world model is bounded by how directly its predictions influence candidate-level scoring. Evaluating each candidate against its own predicted future would allow the planner to exploit action-specific consequences, such as collisions, lane departures, or traffic-rule violations, when distinguishing geometrically similar yet safety-critical trajectories. Realizing this capability requires overcoming two interconnected barriers. At the representation level, predictive features must continue to adapt during planner optimization so that the learned future structure remains aligned with the planning objective. At the planning level, the scorer must evaluate a distinct future latent state for each candidate; shared or pooled futures obscure the action-specific consequences that should guide selection.

To this end, we propose \method{}, a decision-aligned world action model. \method{} maintains predictive supervision throughout planner optimization via a LoRA-adapted Video Joint Embedding Predictive Architecture (V-JEPA)~2.1 online encoder paired with an exponential moving average (EMA) target encoder, ensuring that future representations co-evolve with the driving task rather than being frozen after pretraining. An action-conditioned predictor generates a distinct future latent state for every trajectory candidate through explicit scene--trajectory interaction, and a multi-head trajectory scorer evaluates each candidate jointly with its corresponding latent state. Because offline logs provide observed futures only for the executed expert trajectory, dense JEPA supervision is applied to the expert-matched candidate, while safety-critical hard negatives supply additional local comparisons near planning boundaries. These hard negatives are geometrically similar to the expert-matched candidate but differ in safety outcomes, discouraging the scorer from relying on geometry alone.

Our contributions are threefold:
\begin{itemize}
\item We introduce a compact candidate-conditioned latent-scoring architecture that preserves explicit trajectory--representation pairing while sharing visual scene features across candidates.
\item We develop a joint adaptation scheme in which dense future-feature regression is decoded from the expert-matched compact tokens, directly supervising the matched candidate representation and shared visual encoder alongside planning objectives.
\item We introduce safety-focused trajectory supervision that sharpens decision boundaries between geometrically similar candidates with different safety outcomes, enabling more reliable driving decisions.
\end{itemize}

\section{Related Work}

\subsection{Joint-Embedding Prediction and Latent Planning}

JEPAs capture high-level semantic and temporal representations by predicting latent features of future states rather than reconstructing pixel-level details~\cite{assran2025vjepa2,murlabadia2026vjepa21}. In end-to-end autonomous driving, this paradigm offers an efficient avenue to model scene evolution, which is crucial for forecasting the downstream impact of ego decisions. Recent works such as DriveWorld~\cite{min2024driveworld} and LAW~\cite{li2025enhancing} use future dynamics to enrich visual scene representations for motion planning. Drive-JEPA~\cite{wang2026drivejepa} adapts pretrained video JEPAs to trajectory planning via fine-tuning, while Auto-JEPA~\cite{yang2026autojepa} predicts continuous intent embeddings with a frozen visual backbone to rank candidate paths. Similarly, Latent-WAM~\cite{wang2026latentwam} jointly trains a spatial encoder against an exponential moving average latent target but discards the dynamics branch at test time.

Existing approaches often rely on frozen pretraining, multistage pipelines, or inference-time removal, weakening the coupling between predictive representations and trajectory scoring. In contrast, \method{} jointly adapts the shared encoder with predictive and planning objectives, then refines its compact candidate-future branch and scorer for safety-critical discrimination.

\subsection{Action-Conditioned Driving World Models}

Recognizing that visual environments evolve differently under different ego maneuvers, recent world models explicitly condition future predictions on hypothetical actions or trajectory proposals. Drive-WM~\cite{wang2024driving} synthesizes multiview video futures under alternative driving commands, whereas LAW~\cite{li2025enhancing} and WoTE~\cite{li2025end} predict trajectory-conditioned latent states or bird's-eye-view (BEV) dynamics. World4Drive~\cite{zheng2025world4drive} forecasts multiple intention-guided latent futures and validates paths using an internal evaluator. Concurrently, DriveFuture~\cite{hong2026drivefuture}, IDOL~\cite{zhang2026idol}, and LCDrive~\cite{tan2026latentcot} use future latents for diffusion guidance, inverse-dynamics refinement, and latent chain-of-thought reasoning, respectively.

These methods often aggregate or weakly fuse the predicted futures across proposals, which dilutes the safety consequences specific to each trajectory. In contrast, \method{} feeds a distinct compact future representation for each candidate directly into the scorer, establishing an explicit one-to-one correspondence between prediction and action.

\subsection{Candidate Generation and Trajectory Scoring}
Candidate-based planners generate multiple plausible trajectories and select the optimal one according to scene context and planning objectives. DiffusionDrive efficiently models multimodal action distributions through anchor-guided truncated diffusion~\cite{liao2025diffusiondrive}. DrivoR compresses multi-camera features into camera-aware register tokens and uses separate transformer decoders to generate and score candidates~\cite{kirby2026drivor}. GTRS and ZTRS formulate candidate evaluation as explicit trajectory scoring~\cite{li2025generalized,li2025ztrs}. Beyond geometric and scene-based scoring, DriveSuprim employs coarse-to-fine trajectory selection to distinguish hard-negative trajectories~\cite{yao2026drivesuprim}, and BeyondDrive constructs safety-critical, expert-proximate hard negatives to learn safety boundaries in trajectory space~\cite{wang2026beyonddrive}.

Existing candidate scorers remain largely anchored in current scene geometry and rarely incorporate predicted futures on a per-candidate basis. We augment this paradigm with compact action-conditioned future representations to better distinguish geometrically similar candidates with different safety outcomes.

\section{Method}
\label{sec:method}

\begin{figure}[t]
    \centering
    \methodfigure
    {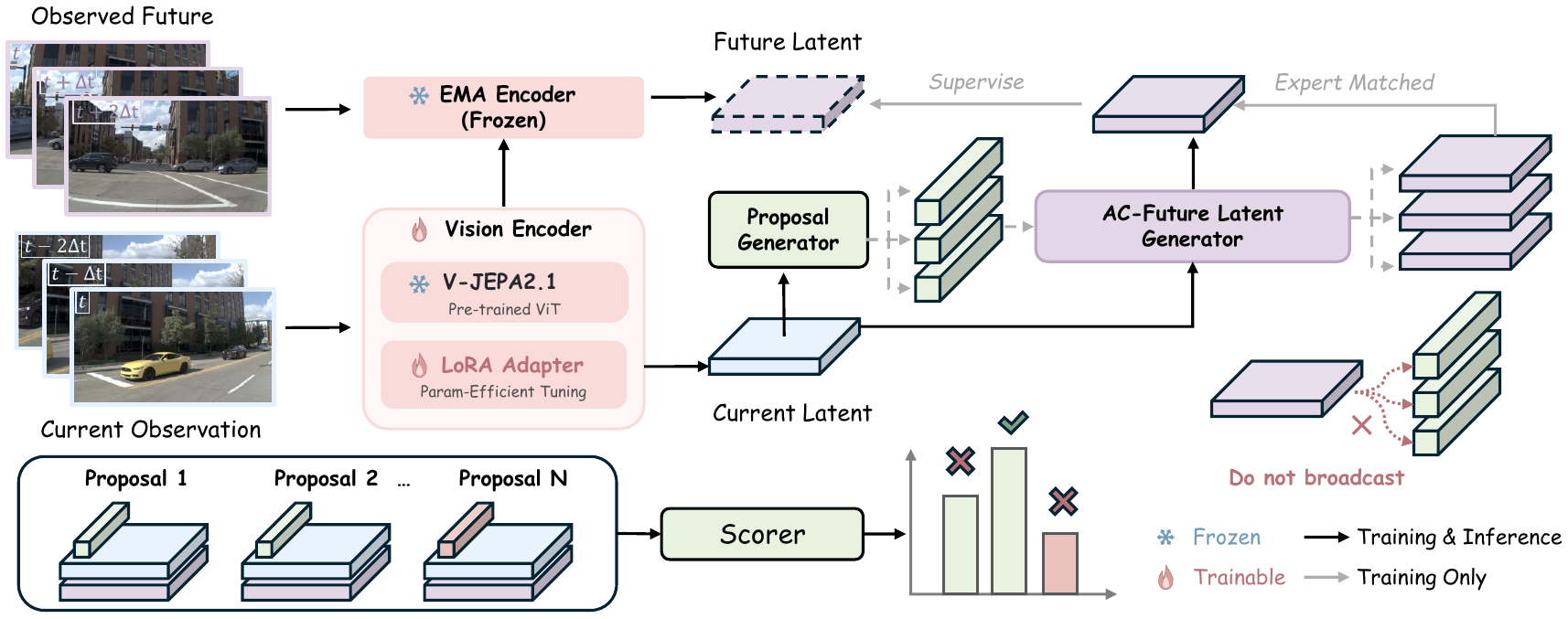}
    {0.98\textwidth}
    {5.4cm}
    \caption{\textbf{Overview of \method{}.} The online visual encoder produces shared scene tokens. A shared action-conditioned predictor constructs compact tokens for each candidate trajectory, and the scorer evaluates each candidate together with its associated tokens and scene context. A training-only dense decoder and a stop-gradient EMA target provide future-feature supervision for the expert-matched candidate.}
    \label{fig:method_overview}
\end{figure}

Given visual observations $X_t$ and generated trajectory candidates $\mathcal T=\{\tau_i\}_{i=1}^{N}$, \method{} predicts candidate utilities and makes the decision. The model has a shared scene encoder, a compact candidate-conditioned scoring path, and a training-only dense future-feature decoder (Fig.~\ref{fig:method_overview}). We describe the inference path first, followed by the predictive and safety-focused supervision used to train it. The candidate count and training settings are reported in Appendix~\ref{app:implementation}; a layer-wise specification and consolidated symbol table are provided in Appendices~\ref{app:compact_network} and~\ref{app:notation}.

\subsection{Candidate-Conditioned Trajectory Scoring}
\label{sec:trajectory_scoring}
The visual encoder and trajectory encoder produce scene tokens $Z_t$ and candidate embeddings $a_i$. A shared compact branch uses each candidate as an attention query over the scene:
\begin{equation}
\begin{aligned}
Z_t &= E_\theta(X_t), \qquad a_i=E_\tau(\tau_i),\\
\widehat z_i &= P_\phi(Q=a_i,K=Z_t,V=Z_t).
\end{aligned}
\label{eq:candidate_future_prediction}
\end{equation}
Here $Z_t\in\mathbb R^{M\times D}$, and $\widehat z_i$ is the compact candidate-conditioned representation. Sharing $P_\phi$ across candidates makes their representations depend on the trajectory--scene interaction, rather than candidate-specific model parameters.

The scorer combines the current scene, candidate, and associated compact representation:
\begin{equation}
\begin{aligned}
h_i &= S_\psi^{\mathrm{enc}}(Z_t,a_i,\widehat z_i),\\
\widehat{\mathbf q}_i &= S_\psi^{\mathrm{factor}}(h_i),\qquad
\widehat s_i=S_\psi^{\mathrm{utility}}(h_i),\\
\tau^\star &= \arg\max_{\tau_i\in\mathcal T}\widehat s_i.
\end{aligned}
\label{eq:trajectory_selection}
\end{equation}
The factor vector contains no-at-fault collision, drivable area compliance, ego progress, time to collision, and comfort. Factor and utility targets are obtained from the planning metrics described in Appendix~\ref{app:evaluation}. The learned utility determines the NAVSIM-v1 candidate ranking.
The implementation of the shared scorer $S_\psi$ is detailed in Appendix~\ref{app:scorer_formulation}.

\begin{figure}[t]
    \centering
    \methodfigure
    {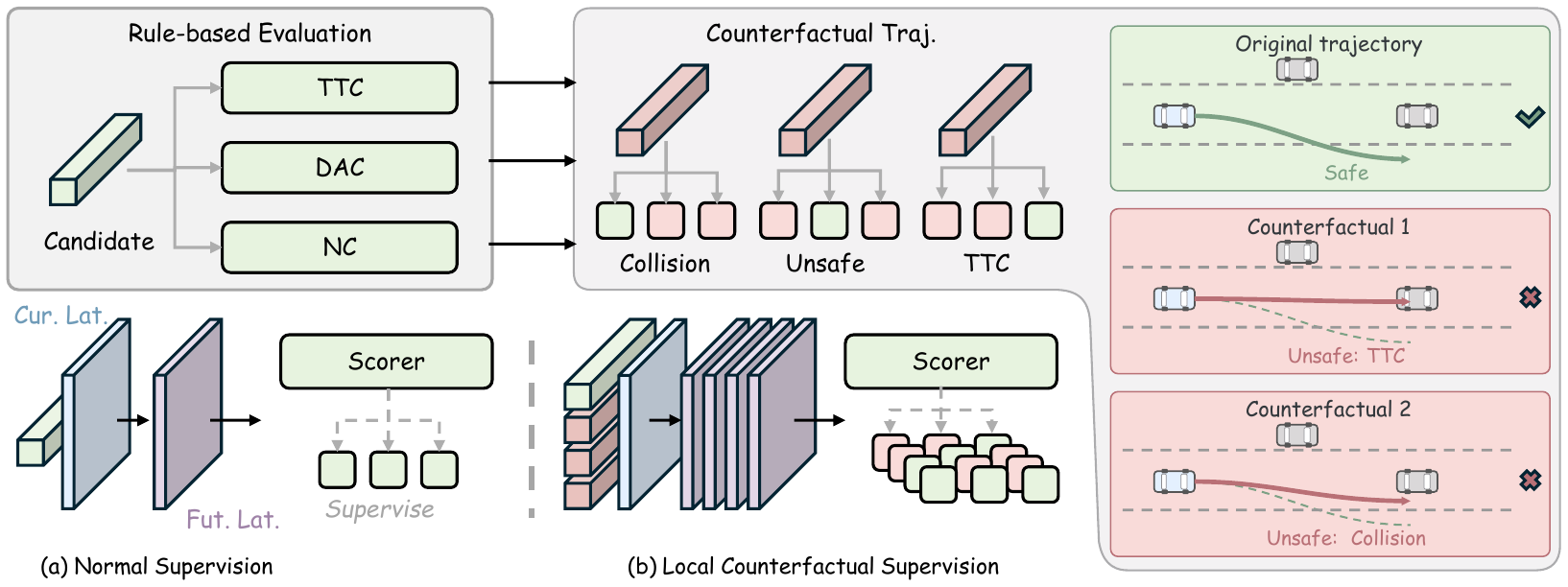}
    {0.88\columnwidth}
    {5.4cm}
    \caption{\textbf{Safety-focused supervision with retrieved trajectory alternatives.} A planning evaluator assigns utility and safety labels to candidate trajectories. Compared with generated-candidate supervision alone (a), the augmented training set (b) includes expert-proximate alternatives with different collision and time-to-collision outcomes. Generated and retrieved trajectories use the same candidate-conditioned predictor and scorer. Retrieved alternatives are supervised by planning labels.}
    \label{fig:hard_anchor_supervision}
\end{figure}

\subsection{Predictive Grounding During Joint Training}

We initialize $E_\theta$ from V-JEPA~2.1 and adapt its query/value projections with LoRA, keeping the original backbone parameters frozen. An EMA target encoder, updated as $\bar\theta\leftarrow\mu\bar\theta+(1-\mu)\theta$, encodes the observed future frame with stop-gradient.

Dense future-feature regression uses the generated candidate closest to the executed expert trajectory, $i^{\mathrm{exp}}=\arg\min_i\operatorname{ADE}(\tau_i,\tau^{\mathrm{exp}})$. ADE is the mean Euclidean ($\ell_2$) displacement over planar positions $(x,y)$. The observed future target is assigned to this candidate in every scene, without an ADE rejection threshold. A dense decoder predicts the matched future token grid:
\begin{equation}
\begin{aligned}
\widetilde Z_{i^{\mathrm{exp}}}&=D_\omega(Z_t,\widehat z_{i^{\mathrm{exp}}}),\qquad
Z_{t+\Delta}=\operatorname{sg}\!\left(E_{\bar\theta}(X_{t+\Delta})\right),\\
\mathcal L_{\mathrm{pred}}&=\frac{1}{MD}\left\|
\operatorname{LN}_c(\widetilde Z_{i^{\mathrm{exp}}})
-\operatorname{LN}_c(Z_{t+\Delta})
\right\|_1.
\end{aligned}
\label{eq:pred}
\end{equation}
Here $\operatorname{LN}_c$ normalizes the channels independently at each spatial position, and the loss is averaged over all $M$ positions and $D$ channels. The dense output $\widetilde Z$ is decoded from the expert-matched compact representation $\widehat z_{i^{\mathrm{exp}}}$, while the scorer uses $\widehat z_i$. During joint training, the prediction loss updates the matched compact tokens, compact predictor, shared encoder adapters, and proposal generator through its native pose tokens. Planning losses supervise all generated and retrieved candidates. Appendix~\ref{app:future_reconstruction_formulation} details the gradient paths.

\subsection{Safety-Focused Supervision and Training}

To emphasize local safety differences, we augment generated candidates with external trajectories, including expert-proximate hard negatives. As illustrated in Fig.~\ref{fig:hard_anchor_supervision}, the hard-negative subset remains geometrically close to the expert while exhibiting worse safety outcomes:
\begin{equation}
d_{\mathrm{traj}}(\tau_j^-,\tau^{\mathrm{exp}})<\epsilon_{\mathrm{geo}},
\qquad
\Delta_{\mathrm{safety}}(\tau_j^-,\tau^{\mathrm{exp}})>\epsilon_{\mathrm{safety}}.
\end{equation}
Each retrieved trajectory passes through the shared compact branch and scorer, with planning-derived factor and utility targets and, in Stage~2, pairwise safety-ranking supervision. Dense future-feature regression uses the expert-matched generated candidate. Appendix~\ref{app:implementation} specifies the sampling distributions and thresholds.

\paragraph{Stage 1: joint adaptation.}
We jointly train the encoder adapters, dense decoder, compact branch, and planner using
\begin{equation}
\mathcal L_1=
\lambda_{\mathrm{pred}}
\mathcal{L}_{\mathrm{pred}}
+
\lambda_{\mathrm{factor}}
\mathcal{L}_{\mathrm{factor}}
+
\lambda_{\mathrm{score}}
\mathcal{L}_{\mathrm{score}}.
\label{eq:total}
\end{equation}
The two planning terms supervise individual driving factors and overall utility, respectively. Generated candidates and external trajectories contribute both targets; the external-trajectory objective is introduced after an initial warm-up. Stage~2 adds pairwise safety ranking.

\paragraph{Stage 2: safety-focused refinement.}
We freeze the visual encoder and its adapters, proposal generator, dense decoder, and EMA target, and optimize the compact branch and scorer. The dense reconstruction loss has zero weight. Stage~2 supervises the five planning factors and learned utility for generated and external candidates. External sampling emphasizes safety-critical cases, unsafe-candidate utility losses receive larger weights, and a safety-ranking objective emphasizes safe--unsafe preferences.

\section{Experiments}
\label{sec:experiments}

\paragraph{Benchmarks and Metrics.}
For NAVSIM, No-at-Fault Collision (NC) measures avoidance of ego-caused collisions, Drivable Area Compliance (DAC) measures whether the trajectory remains within the drivable region, Ego Progress (EP) measures route-aligned advancement, Time to Collision (TTC) captures near-term collision risk, and Comfort evaluates motion smoothness; PDMS aggregates these factors. NAVSIM-v2 additionally reports Driving Direction Compliance (DDC), Traffic Light Compliance (TLC), Lane Keeping (LK), History Comfort (HC), and Extended Comfort (EC), which respectively evaluate correct travel direction, red-light compliance, lane adherence, continuity with the observed motion history, and consistency across consecutive predictions; EPDMS is their extended aggregate score. For Bench2Drive, Avg.~L2 measures trajectory displacement error, Efficiency reflects driving progress and efficiency, Comfortness measures ride smoothness, Success Rate is the fraction of successfully completed routes, and Driving Score summarizes overall closed-loop driving performance. Full implementation details, dataset protocols, and metric definitions are provided in Appendix~\ref{app:evaluation}.

\begin{table}[h]
    \centering
    \caption{\textbf{NAVSIM-v1 benchmark comparison.} }
    \label{tab:benchmark_navsim_v1}
    \resizebox{\mainTableWidth}{!}{%
    \begin{tabular}{lccccccc}
        \toprule
        Method & Venue & NC & DAC & TTC & Comfort & EP & \textbf{PDMS} \\
        \midrule
        PDM-Closed~\cite{dauner2023pdm} & CoRL'23
        & 94.6 & 99.8 & 89.9 & 86.9 & 99.9 & 89.1 \\
        Human driver~\cite{dauner2024navsim} & NeurIPS'24
        & 100.0 & 100.0 & 100.0 & 99.9 & 87.5 & 94.8 \\

        \midrule
        Ego-stat. MLP~\cite{dauner2024navsim} & NeurIPS'24
        & 93.0 & 77.3 & 83.6 & 100.0 & 62.8 & 65.6 \\
        UniVLA~\cite{wang2025unified} & ICLR'26
        & 96.9 & 91.1 & 91.7 & 96.7 & 76.8 & 81.7 \\
        VAD-v2~\cite{jiang2026vadv2} & ICLR'26
        & 98.1 & 94.8 & 94.3 & 100.0 & 80.6 & 86.2 \\
        PRIX~\cite{wozniak2025prix} & RA-L'26
        & 98.1 & 96.3 & 94.1 & 100.0 & 82.3 & 87.8 \\
        DIVER~\cite{song2025breaking} & TPAMI'26
        & 98.5 & 96.5 & 94.9 & 100.0 & 82.6 & 88.3 \\
        DriveVLA-W0~\cite{li2025drivevla} & ICLR'26
        & 98.7 & 99.1 & 95.3 & 99.3 & 83.3 & 90.2 \\
        ReCogDrive~\cite{xiong2026recogdrive} & ICLR'26
        & 97.9 & 97.3 & 94.9 & 100.0 & 87.3 & 90.8 \\
        iPad~\cite{guo2025ipad} & RAL'26
        & 98.6 & 98.3 & 94.9 & 100.0 & 88.0 & 91.7 \\
        SparseDriveV2~\cite{sun2026sparsedrivev2} & ECCV'26
        & 98.5 & 98.4 & 95.0 & 99.9 & 88.6 & 92.0 \\
        DrivoR~\cite{kirby2026drivor} & CVPR'26
        & 98.9 & 98.3 & 96.2 & 100.0 & 89.1 & 93.1 \\
        DriveSuprim~\cite{yao2026drivesuprim} & AAAI'26
        & 98.6 & 98.6 & 95.5 & 100.0 & 91.3 & 93.5 \\

        \midrule
        \highlightrow
        \method{} & --
        & 99.1 & 98.9 & 96.8 & 99.8 & 90.0 & \textbf{93.7} \\
        \bottomrule
    \end{tabular}
    }
\end{table}

\subsection{Main Results}

Tables~\ref{tab:benchmark_navsim_v1}--\ref{tab:benchmark_bench2drive} summarize the main results. \method{} achieves 93.7 PDMS on NAVSIM-v1 using its learned candidate scores directly, surpassing prior learned planners while attaining 99.1 NC, 98.9 DAC, and 90.0 EP. Training-free temporal-consistency reranking yields 89.22 EPDMS on NAVSIM-v2, exceeding the strongest comparison by 1.72 points. On Bench2Drive, \method{} obtains the highest Driving Score of 59.04 with a 24.66\% Success Rate. Notably, while \method{} does not lead in progress-oriented metrics, it excels in safety-critical metrics such as NC, TTC, and DAC, highlighting its strength in prioritizing safe planning over mere forward advancement.

\begin{table}[h]
    \centering
    \caption{\textbf{NAVSIM-v2 benchmark comparison.}}
    \label{tab:benchmark_navsim_v2}
    \resizebox{\mainTableWidth}{!}{%
    \begin{tabular}{lccccccccccc}
        \toprule
        Method & Img. Backbone
        & NC & DAC & DDC & TLC
        & EP & TTC & LK & HC
        & EC & EPDMS \\
        \midrule
        Ego Status MLP & ResNet-34
        & 93.1 & 77.9 & 92.7 & 99.6 & 86.0 & 91.5 & 89.4 & 98.3 & 85.4
        & 64.0 \\
        TransFuser~\cite{chitta2022transfuser} & ResNet-34
        & 96.9 & 89.9 & 97.8 & 99.7 & 87.1 & 95.4 & 92.7 & 98.3 & 87.2
        & 76.7 \\
        Hydra-MDP++~\cite{li2025hydramdppp} & ResNet-34
        & 97.2 & 97.5 & 99.4 & 99.6 & 83.1 & 96.5 & 94.4 & 98.2 & 70.9
        & 81.4 \\
        DriveSuprim~\cite{yao2026drivesuprim} & ResNet-34
        & 97.5 & 96.5 & 99.4 & 99.6 & 88.4 & 96.6 & 95.5 & 98.3 & 77.0
        & 83.1 \\
        ARTEMIS~\cite{feng2025artemis} & ResNet-34
        & 98.3 & 95.1 & 98.6 & 99.8 & 81.5 & 97.4 & 96.5 & 98.3 & 98.3
        & 83.1 \\
        DiffusionDriveV2~\cite{zou2025diffusiondrivev2} & ResNet-34
        & 97.7 & 96.6 & 99.2 & 99.8 & 88.9 & 97.2 & 96.0 & 97.8 & 91.0
        & 87.5 \\
        SparseDriveV2~\cite{sun2026sparsedrivev2} & ResNet-34
        & 98.1 & 98.1 & 99.6 & 99.8 & 91.1 & 97.3 & 96.9 & 98.2 & 78.4
        & 86.7 \\
        \midrule
        Hydra-MDP++~\cite{li2025hydramdppp} & ViT/L
        & 98.4 & 98.0 & 99.4 & 99.8 & 87.5 & 97.7 & 95.3 & 98.3 & 77.4
        & 85.1 \\
        DriveSuprim~\cite{yao2026drivesuprim} & ViT/L
        & 97.8 & 97.9 & 99.5 & 99.9 & 90.6 & 97.1 & 96.6 & 98.3 & 77.9
        & 86.0 \\
        \midrule
        \highlightrow
        \method{} & ViT/L
& 99.2 & 98.8 & 97.8 & 99.5
& 91.1 & 98.7 & 93.3 & 96.7
& 77.1 & \textbf{89.2}  \\
        \bottomrule
    \end{tabular}%
    }
\end{table}

\begin{table}[t]
    \centering
    \small
    \setlength{\tabcolsep}{4.0pt}
    \caption{\textbf{Bench2Drive closed-loop comparison.} }
    \label{tab:benchmark_bench2drive}
    \resizebox{\mainTableWidth}{!}{%
    \begin{tabular}{lccccc}
        \toprule
        Method
        & Avg. L2 (m) $\downarrow$
        & Efficiency $\uparrow$
        & Comfortness $\uparrow$
        & Success Rate (\%) $\uparrow$
        & Driving Score $\uparrow$ \\
        \midrule
        TCP-traj w/o distillation~\cite{wu2022tcp}
        & 1.96 & 78.78 & 22.96 & 20.45 & 49.30 \\
        AD-MLP~\cite{zhai2023admlp}
        & 3.64 & 48.45 & 22.63 & 0.00 & 18.05 \\
        UniAD-Tiny~\cite{hu2023uniad}
        & 0.80 & 123.92 & 47.04 & 13.18 & 40.73 \\
        UniAD-Base~\cite{hu2023uniad}
        & 0.73 & 129.21 & 43.58 & 16.36 & 45.81 \\
        VAD~\cite{jiang2023vad}
        & 0.91 & 157.94 & 46.01 & 15.00 & 42.35 \\
        GenAD~\cite{zheng2024genad}
        & -- & -- & -- & 15.90 & 44.81 \\
        EgoFSD-S~\cite{su2024egofsd}
        & 0.70 & 178.30 & -- & 21.00 & 52.02 \\
        ReAL-AD~\cite{lu2025realad}
        & 0.84 & -- & -- & 11.36 & 41.17 \\
        BridgeAD~\cite{zhang2025bridging}
        & 0.71 & -- & -- & 22.73 & 50.06 \\
        SeerDrive~\cite{zhang2026future}
        & 0.66 & -- & -- & 30.17 & 58.32 \\
        \midrule
        \highlightrow
        \method{}
        & 0.95 & 140.40 & 14.64 & 24.66 & \textbf{59.04} \\
        \bottomrule
    \end{tabular}%
    }
\end{table}

\subsection{Ablation Studies}
\label{sec:ablations}

\begin{figure}[tbp]
    \centering
    \methodfigure
    {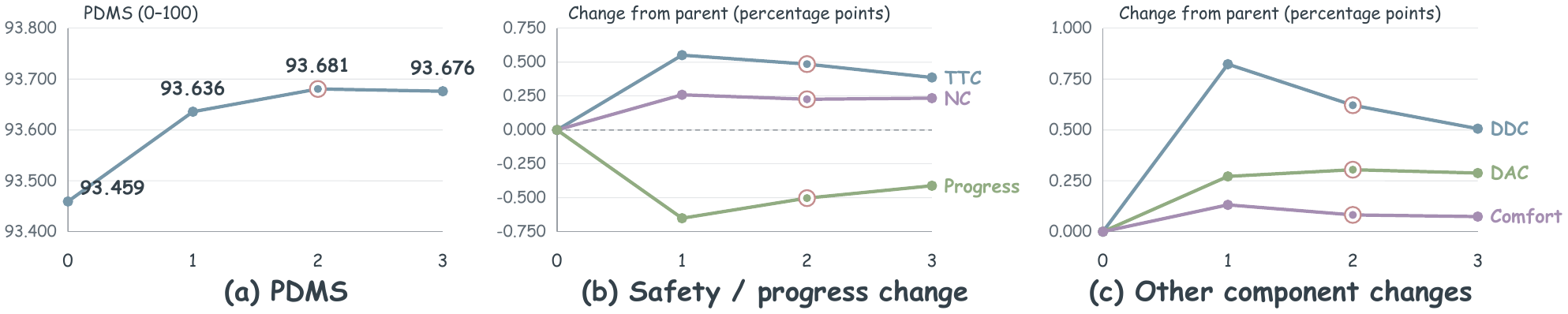}
    {0.98\textwidth}
    {4.2cm}
    \caption{\textbf{Safety--progress trade-offs during refinement.} Epoch 0 denotes Stage~1, and epochs 1--3 denote completed refinement epochs; circles mark the retained checkpoint.}
    \label{fig:stage2_checkpoint_analysis}
\end{figure}

\paragraph{Candidate Conditioning and Predictive Supervision.}
Table~\ref{tab:main_results} isolates the effects of future conditioning and Stage~2 safety-focused supervision. Before Stage~2, the action-conditioned future achieves the highest PDMS of 93.46 among the conditioning variants, outperforming no future prediction (93.31), shared global future (92.81), and current-latent conditioning (93.25). Stage~2 further increases PDMS to 93.68 and improves NC, DAC, TTC, and Comfort, despite a 0.50-point reduction in EP. 
Fig.~\ref{fig:stage2_checkpoint_analysis} tracks the evolution of each metric throughout Stage~2. The overall improvement is driven primarily by gains in safety-related metrics, while PDMS approaches a plateau after two completed refinement epochs.
The improvement indicates that candidate-conditioned future latents provide decision-relevant information for more effective trajectory selection.
Table~\ref{tab:encoder_dense_ablation} evaluates the predictive-representation design. The settings without and with dense loss correspond to the V-JEPA~2.0 and V-JEPA~2.1 objectives, respectively. Adding dense loss improves PDMS from 91.26 to 91.95 with a frozen encoder and from 92.74 to 92.98 with LoRA adaptation. With the dense objective enabled, LoRA achieves 92.98 PDMS, outperforming both full fine-tuning (92.62) and the frozen encoder (91.95). Because LoRA keeps the pretrained backbone parameters fixed and updates only low-rank adapters, it preserves the original encoder while adapting its representations to the driving task. Appendix Fig.~\ref{fig:lora_feature_comparison} further shows that LoRA redistributes attention and reorganizes features around decision-relevant vehicles and traffic signals while retaining broader scene context. Finally, under LoRA adaptation with dense supervision, the EMA target achieves the best PDMS of 93.68, supporting the effectiveness of momentum-based target updates.

\begin{table}[tbp]
    \centering
    \small
    
    \caption{\textbf{Ablation on action-conditioned futures and safety-focused refinement.}}
    \label{tab:main_results}
    \resizebox{\mainTableWidth}{!}{%
    \begin{tabular}{lcrrrrrr}
        \toprule
        Configuration
        & Safety sup.
        & PDMS
        & NC
        & DAC
        & EP
        & TTC
        & Comfort \\
        \midrule
        No Future Prediction
        & \xmark
        & 93.31
        & 98.45
        & 98.27
        & 91.36
        & 95.48
        & \textbf{99.99} \\
        Shared Global Future
        & \xmark
        & 92.81
        & 99.02
        & 98.46
        & 88.68
        & 96.54
        & \textbf{99.99} \\
        Current-Latent Conditioning
        & \xmark
        & 93.25
        & 98.44
        & 98.19
        & \textbf{91.38}
        & 95.49
        & 99.94 \\
        \multirow{2}{*}{Action-Conditioned Future}
        & \xmark
        & 93.46
        & 98.88
        & 98.58
        & 90.47
        & 96.33
        & 99.69 \\
        \cmidrule(lr){2-8}
        \highlightrow
        & \cmark
        & \textbf{93.68}
        & \textbf{99.11}
        & \textbf{98.88}
        & 89.97
        & \textbf{96.81}
        & 99.77 \\
        \bottomrule
    \end{tabular}
    }
\end{table}

\begin{table}[tbp]
    \centering
    \captionsetup{
        font=small,
        labelfont=bf,
        justification=raggedright,
        singlelinecheck=false,
        position=top,
        skip=5pt
    }
    \footnotesize
    \renewcommand{\arraystretch}{1.20}
    \setlength{\tabcolsep}{3pt}
    \setlength{\heavyrulewidth}{0.7pt}
    \setlength{\lightrulewidth}{0.4pt}
    \newcommand{\predrepcolumnwidth}{0.56\linewidth}
    \newcommand{\parametercolumnwidth}{0.42\linewidth}
    \newcommand{\ablationcell}[1]{%
        \cellcolor{ablationBlue}\textbf{#1}%
    }
    \settowidth{\ablationLabelWidth}{%
        \textbf{Horizon $\Delta$ (s)}%
    }
    
    \resizebox{\mainTableWidth}{!}{%
    \begin{minipage}{\textwidth}
    \centering
    \begin{minipage}[t]{\predrepcolumnwidth}
        \vspace{0pt}

        \caption{Ablation of supervision, encoder, and target.}
        \label{tab:encoder_dense_ablation}

        \centering
        \begin{tabularx}{\linewidth}{
            l
            >{\centering\arraybackslash}X
            >{\centering\arraybackslash}X
            r
        }
            \toprule
            \textbf{Adaptation}
                & \textbf{Dense loss}
                & \textbf{Target}
                & \textbf{PDMS} \\
            \midrule

            \multicolumn{4}{l}{%
                \color{ablationGray}\itshape
                Online encoder and predictive objective
            } \\
            \addlinespace[2pt]

            Frozen   & \xmark & Frozen & 91.26 \\
            Frozen   & \cmark & Frozen & 91.95 \\
            LoRA     & \xmark & Frozen & 92.74 \\
            LoRA     & \cmark & Frozen & 92.98 \\
            Full ft. & \cmark & Frozen & 92.62 \\

            \midrule
            \multicolumn{4}{l}{%
                \color{ablationGray}\itshape
                Target policy (LoRA + dense loss)
            } \\
            \addlinespace[2pt]

            LoRA & \cmark & Separate & 93.10 \\
            LoRA & \cmark & Shared   & 93.34 \\

            \rowcolor{ablationBlue}
            LoRA & \cmark & \textbf{EMA} & \textbf{93.68} \\

            \bottomrule
        \end{tabularx}
    \end{minipage}\hfill
    \begin{minipage}[t]{\parametercolumnwidth}
        \vspace{0pt}
        \setlength{\tabcolsep}{1pt}
        \caption{Ablation on horizon (Stage~1).}
        \label{tab:prediction_horizon_ablation}

        \centering
        \begin{tabularx}{\linewidth}{
            p{\ablationLabelWidth}
            *{5}{>{\centering\arraybackslash}X}
        }
            \toprule
            \textbf{Horizon $\Delta$ (s)}
                & 0.0
                & \ablationcell{0.5}
                & 1.0
                & 1.5
                & 2.0 \\
            \midrule
            PDMS
                & 93.25
                & \ablationcell{93.46}
                & 93.37
                & 93.23
                & 92.79 \\
            \bottomrule
        \end{tabularx}

        \vspace{5pt}
        \caption{Ablation on LoRA rank (Stage~1).}
        \label{tab:lora_rank_ablation}

        \begin{tabularx}{\linewidth}{
            p{\ablationLabelWidth}
            *{5}{>{\centering\arraybackslash}X}
        }
            \toprule
            \textbf{LoRA rank $r$}
                & 0
                & 8
                & 16
                & \ablationcell{32}
                & 64 \\
            \midrule
            PDMS
                & 91.95
                & 93.10
                & 93.21
                & \ablationcell{93.46}
                & 93.08 \\
            \bottomrule
        \end{tabularx}

        \vspace{5pt}
        \caption{Ablation on candidate count.}
        \label{tab:candidate_count_ablation}

        \begin{tabularx}{\linewidth}{
            p{\ablationLabelWidth}
            *{5}{>{\centering\arraybackslash}X}
        }
            \toprule
            \textbf{Candidates}
                & 1
                & 8
                & 16
                & \ablationcell{32}
                & 64 \\
            \midrule
            PDMS
                & 87.11
                & 90.76
                & 91.89
                & \ablationcell{93.68}
                & \textbf{93.68} \\
            \bottomrule
        \end{tabularx}
    \end{minipage}
    \end{minipage}%
    }
\end{table}

\paragraph{Sensitivity to Prediction Horizon, Adapter Rank, and Candidate Count.}
Tables~\ref{tab:prediction_horizon_ablation}--\ref{tab:candidate_count_ablation} study the prediction horizon, LoRA rank, and candidate count, with the first two reporting Stage~1 performance. Each horizon uses a separately trained checkpoint; $\Delta=0$ predicts the current latent state, while $\Delta=0.5$ achieves the best PDMS of 93.46 and is used for quantitative evaluation. The 1.5-second model is used only for visualization. LoRA rank 32 performs best, while increasing the rank to 64 degrades performance. PDMS increases from 87.11 with one candidate to 93.68 with 32 candidates and remains unchanged with 64, motivating 32 as the default. Appendix~\ref{app:forward_time} reports the corresponding inference-time scaling.

\subsection{Qualitative Analysis}

\begin{figure}[!htbp]
    \centering
    \methodfigure
    {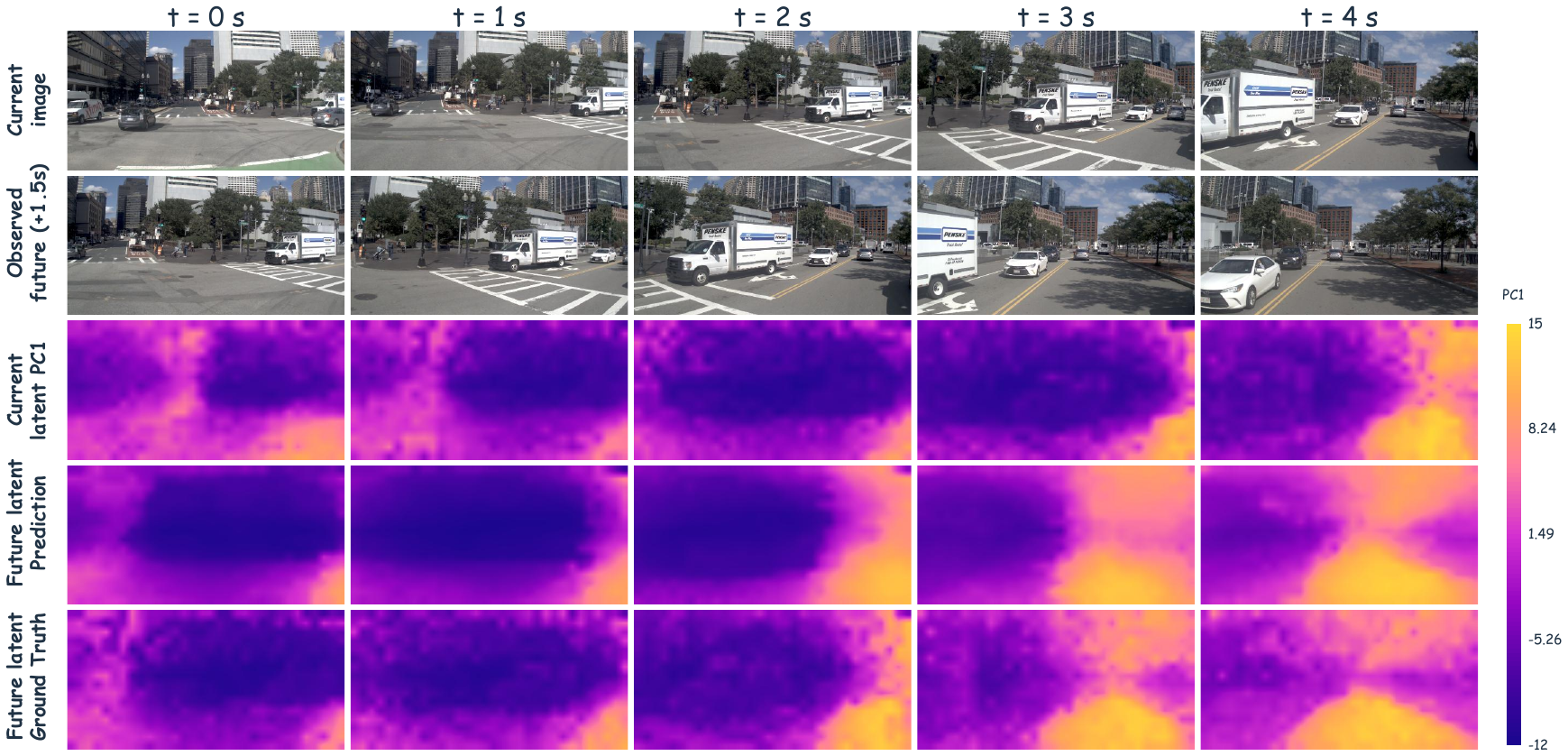}
    {0.90\textwidth}
    {7.0cm}
    \caption{\textbf{Future-feature evolution in a truck scene.} Predictions from the 1.5-second visualization checkpoint. Each column shows an independent prediction.}
    \label{fig:truck_future_latent}
\end{figure}

\paragraph{Future-Feature Evolution.}
Fig.~\ref{fig:truck_future_latent} visualizes future-feature prediction in a dynamic truck scene. The predictions capture broad spatial evolution relevant to decision making. In particular, the high-valued yellow regions increasingly align with the open roadway visible in the future frames, suggesting that the framework emphasizes large-scale free-space and scene-layout changes relevant to trajectory evaluation. Additional future-feature visualizations are provided in Appendix Figs.~\ref{fig:crossing_future_latent} and~\ref{fig:suv_future_latent}.

\begin{figure}[tbp]
    \centering
    \methodfigure
    {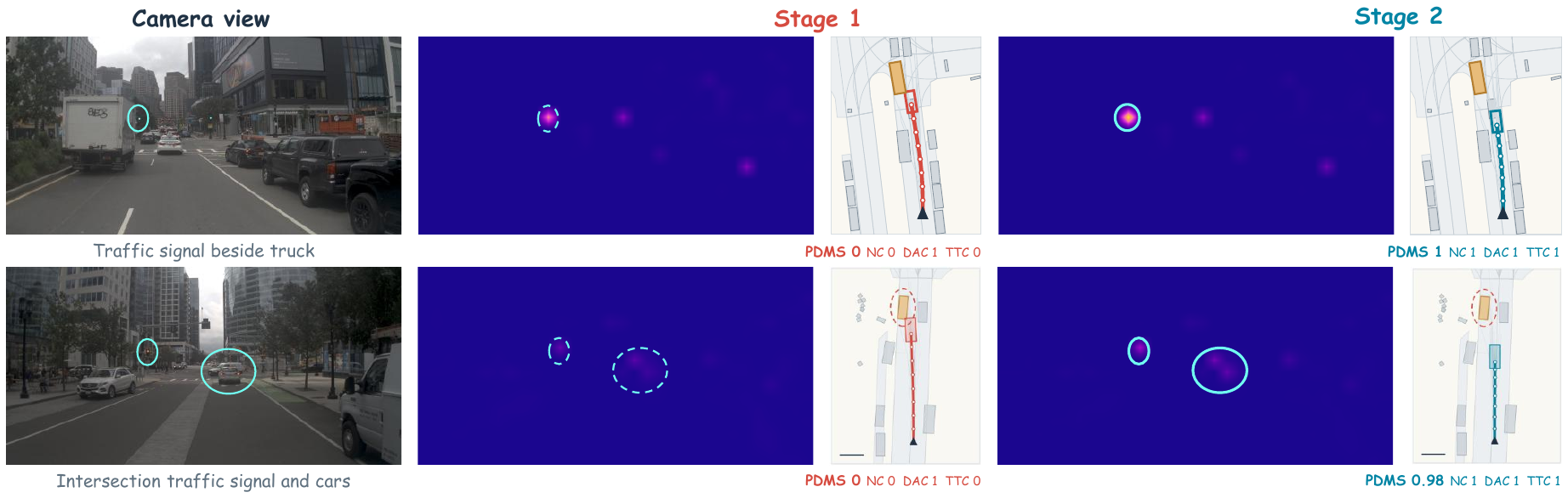}
    {0.90\textwidth}
    {4.0cm}
    \caption{\textbf{Safety-critical attention and safer decision making.}}
    \label{fig:stage2_attention}
\end{figure}

\paragraph{Safety-Focused Candidate Refinement.}
Fig.~\ref{fig:stage2_attention} illustrates the source of the Stage~2 improvement. The first scene features a traffic signal adjacent to a large truck in dense traffic, while the second depicts an intersection signal with multiple interacting vehicles. After refinement, the attention responses become more concentrated on the annotated traffic signals and relevant vehicles. Correspondingly, Stage~2 replaces the collision-prone Stage~1 trajectories with safer alternatives, thereby resolving the NC and TTC failures and increasing PDMS from zero to 1.00 and 0.98, respectively. These examples indicate that safety-focused refinement enhances candidate discrimination by better exploiting safety-critical scene cues.

\begin{figure}[tbp]
    \centering
    \includegraphics[width=0.95\textwidth]{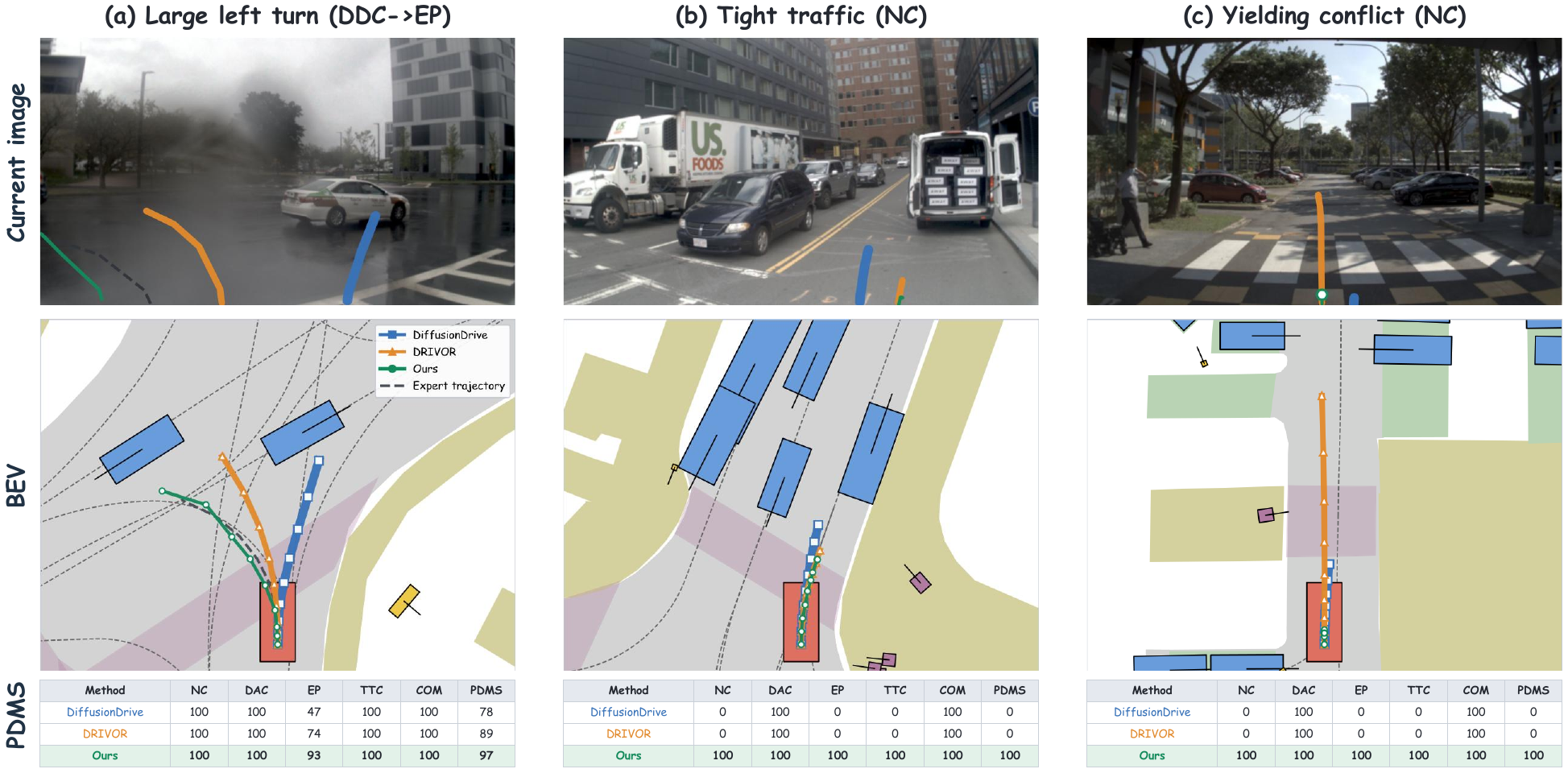}
    \caption{\textbf{Direction Following and Safety-Critical Interaction.} Camera views, bird's-eye-view trajectories, and per-scene metrics compare DiffusionDrive, DrivoR, and \method{} in a large left turn, tight traffic, and a yielding conflict.}
    \label{fig:qualitative_results}
\end{figure}

\paragraph{Direction Following and Safety-Critical Interaction.}
Fig.~\ref{fig:qualitative_results} compares \method{} with DiffusionDrive and DrivoR in three representative scenarios. In (a), \method{} follows the intended left-turn direction and expert route more closely, achieving higher progress while maintaining safe behavior. In the tight-traffic scene (b), it responds more appropriately to the leading vehicle and avoids the collision risks encountered by both baselines. In the yielding-conflict scene (c), \method{} better accounts for the traffic-control cue and crossing conflict, selecting a safe trajectory instead of proceeding into the conflict. Together, these examples show improved direction adherence, interaction with surrounding vehicles, and compliance with traffic controls.

\section{Conclusion}

We presented \method{}, a decision-aligned world action model that integrates compact action-conditioned future representations with candidate-level trajectory scoring.
The method combines joint predictive and planning adaptation with safety-focused post-training to align candidate representations with trajectory evaluation.
Experiments on NAVSIM-v1, NAVSIM-v2, and Bench2Drive support the approach, with training-free temporal-consistency reranking used on NAVSIM-v2.
Ablations favor candidate-specific futures over no-future, current-latent, and shared-future controls, and support LoRA adaptation with dense predictive supervision and an EMA target.
Sensitivity studies examine prediction horizon, adapter rank, and candidate count, while qualitative diagnostics characterize representation reorganization and coarse future-feature evolution.
Post-training analysis further shows that improved safety and compliance come with a modest reduction in progress.
Overall, these findings highlight the value of aligning future representations with candidate evaluation rather than treating prediction as an isolated auxiliary task.

\subsection*{AI use statement}

Generative AI tools were used to assist with language editing, \LaTeX{}
formatting, and code inspection. All technical content, experimental designs,
implementations, results, and claims were reviewed and verified by the authors.
The authors take full responsibility for the final content of this work.

\subsection*{Reproducibility statement}

The model formulation and two-stage training objectives are described in
\secref{sec:method}. Benchmark results and controlled ablations are reported
in \secref{sec:experiments}. Appendix~\ref{app:evaluation} defines the datasets,
evaluation protocols, and metrics, while Appendix~\ref{app:implementation}
provides the architecture, optimization, post-training, and inference details.
Project code is available at \url{https://github.com/LeapWM/da-wam}.
Additional representation diagnostics, ablations, and efficiency measurements
are provided in Appendices~\ref{app:representation_mechanism}
and~\ref{app:additional_analysis}.

\FloatBarrier
\bibliography{references}

\begin{thebibliography}{42}
\providecommand{\natexlab}[1]{#1}
\providecommand{\url}[1]{\texttt{#1}}
\expandafter\ifx\csname urlstyle\endcsname\relax
  \providecommand{\doi}[1]{doi: #1}\else
  \providecommand{\doi}{doi: \begingroup \urlstyle{rm}\Url}\fi

\bibitem[Assran et~al.(2025)Assran, Bardes, Fan, Garrido, Howes, Muckley,
  Rizvi, Roberts, Sinha, Zholus, et~al.]{assran2025vjepa2}
Mido Assran, Adrien Bardes, David Fan, Quentin Garrido, Russell Howes, Matthew
  Muckley, Ammar Rizvi, Claire Roberts, Koustuv Sinha, Artem Zholus, et~al.
\newblock V-jepa 2: Self-supervised video models enable understanding,
  prediction and planning.
\newblock \emph{arXiv preprint arXiv:2506.09985}, 2025.

\bibitem[Chitta et~al.(2022)Chitta, Prakash, Jaeger, Yu, Renz, and
  Geiger]{chitta2022transfuser}
Kashyap Chitta, Aditya Prakash, Bernhard Jaeger, Zehao Yu, Katrin Renz, and
  Andreas Geiger.
\newblock Transfuser: Imitation with transformer-based sensor fusion for
  autonomous driving.
\newblock \emph{IEEE transactions on pattern analysis and machine
  intelligence}, 45\penalty0 (11):\penalty0 12878--12895, 2022.

\bibitem[Dauner et~al.(2023)Dauner, Hallgarten, Geiger, and
  Chitta]{dauner2023pdm}
Daniel Dauner, Marcel Hallgarten, Andreas Geiger, and Kashyap Chitta.
\newblock Parting with misconceptions about learning-based vehicle motion
  planning.
\newblock In \emph{Conference on Robot Learning}, pages 1268--1281. PMLR, 2023.

\bibitem[Dauner et~al.(2024)Dauner, Hallgarten, Li, Weng, Huang, Yang, Li,
  Gilitschenski, Ivanovic, Pavone, et~al.]{dauner2024navsim}
Daniel Dauner, Marcel Hallgarten, Tianyu Li, Xinshuo Weng, Zhiyu Huang, Zetong
  Yang, Hongyang Li, Igor Gilitschenski, Boris Ivanovic, Marco Pavone, et~al.
\newblock Navsim: Data-driven non-reactive autonomous vehicle simulation and
  benchmarking.
\newblock \emph{Advances in Neural Information Processing Systems},
  37:\penalty0 28706--28719, 2024.

\bibitem[Feng et~al.(2026)Feng, Xi, Chu, Wang, Deng, Wang, Lu, Wang, and
  Huang]{feng2025artemis}
Renju Feng, Ning Xi, Duanfeng Chu, Rukang Wang, Zejian Deng, Anzheng Wang,
  Liping Lu, Jinxiang Wang, and Yanjun Huang.
\newblock Artemis: Autoregressive end-to-end trajectory planning with mixture
  of experts for autonomous driving.
\newblock \emph{IEEE Robotics and Automation Letters}, 11\penalty0
  (1):\penalty0 226--233, 2026.
\newblock \doi{10.1109/LRA.2025.3632616}.

\bibitem[Guo et~al.(2026)Guo, Liu, Wu, Pan, and Lv]{guo2025ipad}
Ke~Guo, Haochen Liu, Xiaojun Wu, Jia Pan, and Chen Lv.
\newblock ipad: Iterative proposal-centric end-to-end autonomous driving.
\newblock \emph{IEEE Robotics and Automation Letters}, 2026.

\bibitem[Hong et~al.(2026)Hong, Zhou, Li, Zhou, Liu, Luo, Xu, Yang, and
  Song]{hong2026drivefuture}
Yufeng Hong, Xiaotian Zhou, Yingyan Li, Xiangpo Zhou, Lin Liu, Yadan Luo,
  Shaoqing Xu, Lei Yang, and Ziying Song.
\newblock Drivefuture: Future-aware latent world models for autonomous driving.
\newblock \emph{arXiv preprint arXiv:2605.09701}, 2026.

\bibitem[Hu et~al.(2023)Hu, Yang, Chen, Li, Sima, Zhu, Chai, Du, Lin, Wang,
  et~al.]{hu2023uniad}
Yihan Hu, Jiazhi Yang, Li~Chen, Keyu Li, Chonghao Sima, Xizhou Zhu, Siqi Chai,
  Senyao Du, Tianwei Lin, Wenhai Wang, et~al.
\newblock Planning-oriented autonomous driving.
\newblock In \emph{2023 IEEE/CVF Conference on Computer Vision and Pattern
  Recognition (CVPR)}, pages 17853--17862. IEEE, 2023.

\bibitem[Jiang et~al.(2023)Jiang, Chen, Xu, Liao, Chen, Zhou, Zhang, Liu,
  Huang, and Wang]{jiang2023vad}
Bo~Jiang, Shaoyu Chen, Qing Xu, Bencheng Liao, Jiajie Chen, Helong Zhou, Qian
  Zhang, Wenyu Liu, Chang Huang, and Xinggang Wang.
\newblock Vad: Vectorized scene representation for efficient autonomous
  driving.
\newblock In \emph{2023 IEEE/CVF International Conference on Computer Vision
  (ICCV)}, pages 8306--8316. IEEE, 2023.

\bibitem[Jiang et~al.(2026)Jiang, Chen, Gao, Liao, Zhang, Liu, and
  Wang]{jiang2026vadv2}
Bo~Jiang, Shaoyu Chen, Hao Gao, Bencheng Liao, Qian Zhang, Wenyu Liu, and
  Xinggang Wang.
\newblock Vadv2: End-to-end vectorized autonomous driving via probabilistic
  planning.
\newblock In \emph{International Conference on Learning Representations},
  volume 2026, pages 68886--68900, 2026.

\bibitem[Kirby et~al.(2026)Kirby, Boulch, Xu, Yin, Puy, Zablocki, Bursuc,
  Gidaris, Marlet, Bartoccioni, et~al.]{kirby2026drivor}
Ellington Kirby, Alexandre Boulch, Yihong Xu, Yuan Yin, Gilles Puy, {\'E}loi
  Zablocki, Andrei Bursuc, Spyros Gidaris, Renaud Marlet, Florent Bartoccioni,
  et~al.
\newblock Driving on registers.
\newblock In \emph{Proceedings of the IEEE/CVF Conference on Computer Vision
  and Pattern Recognition}, pages 32058--32069, 2026.

\bibitem[Li et~al.(2025{\natexlab{a}})Li, Li, Lan, Xie, Zhang, Liu, Wu, Yu, and
  Alvarez]{li2025hydramdppp}
Kailin Li, Zhenxin Li, Shiyi Lan, Yuan Xie, Zhizhong Zhang, Jiayi Liu, Zuxuan
  Wu, Zhiding Yu, and Jose~M Alvarez.
\newblock Hydra-mdp++: Advancing end-to-end driving via expert-guided
  hydra-distillation.
\newblock \emph{arXiv preprint arXiv:2503.12820}, 2025{\natexlab{a}}.

\bibitem[Li et~al.(2025{\natexlab{b}})Li, Fan, He, Wang, Chen, Zhang, and
  Tan]{li2025enhancing}
Yingyan Li, Lue Fan, Jiawei He, Yuqi Wang, Yuntao Chen, Zhaoxiang Zhang, and
  Tieniu Tan.
\newblock Enhancing end-to-end autonomous driving with latent world model.
\newblock In \emph{International Conference on Learning Representations},
  volume 2025, pages 42942--42959, 2025{\natexlab{b}}.

\bibitem[Li et~al.(2025{\natexlab{c}})Li, Wang, Liu, He, Fan, and
  Zhang]{li2025end}
Yingyan Li, Yuqi Wang, Yang Liu, Jiawei He, Lue Fan, and Zhaoxiang Zhang.
\newblock End-to-end driving with online trajectory evaluation via bev world
  model.
\newblock In \emph{2025 IEEE/CVF International Conference on Computer Vision
  (ICCV)}, pages 27137--27146. IEEE, 2025{\natexlab{c}}.

\bibitem[Li et~al.(2026)Li, Shang, Liu, Zhan, Wang, Wang, Chen, Wang, An, Tang,
  et~al.]{li2025drivevla}
Yingyan Li, Shuyao Shang, Weisong Liu, Bing Zhan, Haochen Wang, Yuqi Wang,
  Yuntao Chen, Xiaoman Wang, Yasong An, Chufeng Tang, et~al.
\newblock Drivevla-w0: World models amplify data scaling law in autonomous
  driving.
\newblock In \emph{International Conference on Learning Representations},
  volume 2026, pages 7890--7911, 2026.

\bibitem[Li et~al.(2025{\natexlab{d}})Li, Yao, Wang, Sun, Chen, Chang, Shen,
  Song, Wu, Lan, et~al.]{li2025ztrs}
Zhenxin Li, Wenhao Yao, Zi~Wang, Xinglong Sun, Jingde Chen, Nadine Chang,
  Maying Shen, Jingyu Song, Zuxuan Wu, Shiyi Lan, et~al.
\newblock Ztrs: Zero-imitation end-to-end autonomous driving with trajectory
  scoring.
\newblock \emph{arXiv preprint arXiv:2510.24108}, 2025{\natexlab{d}}.

\bibitem[Li et~al.(2025{\natexlab{e}})Li, Yao, Wang, Sun, Chen, Chang, Shen,
  Wu, Lan, and Alvarez]{li2025generalized}
Zhenxin Li, Wenhao Yao, Zi~Wang, Xinglong Sun, Joshua Chen, Nadine Chang,
  Maying Shen, Zuxuan Wu, Shiyi Lan, and Jose~M Alvarez.
\newblock Generalized trajectory scoring for end-to-end multimodal planning.
\newblock \emph{arXiv preprint arXiv:2506.06664}, 2025{\natexlab{e}}.

\bibitem[Liao et~al.(2025)Liao, Chen, Yin, Jiang, Wang, Yan, Zhang, Li, Zhang,
  Zhang, et~al.]{liao2025diffusiondrive}
Bencheng Liao, Shaoyu Chen, Haoran Yin, Bo~Jiang, Cheng Wang, Sixu Yan, Xinbang
  Zhang, Xiangyu Li, Ying Zhang, Qian Zhang, et~al.
\newblock Diffusiondrive: Truncated diffusion model for end-to-end autonomous
  driving.
\newblock In \emph{2025 IEEE/CVF Conference on Computer Vision and Pattern
  Recognition (CVPR)}, pages 12037--12047. IEEE, 2025.

\bibitem[Lu et~al.(2025)Lu, Tu, Ma, and Zhu]{lu2025realad}
Yuhang Lu, Jiadong Tu, Yuexin Ma, and Xinge Zhu.
\newblock Real-ad: Towards human-like reasoning in end-to-end autonomous
  driving.
\newblock In \emph{2025 IEEE/CVF International Conference on Computer Vision
  (ICCV)}, pages 27783--27793. IEEE, 2025.

\bibitem[Min et~al.(2024)Min, Zhao, Xiao, Zhao, Xu, Zhu, Jin, Li, Guo, Xing,
  et~al.]{min2024driveworld}
Chen Min, Dawei Zhao, Liang Xiao, Jian Zhao, Xinli Xu, Zheng Zhu, Lei Jin,
  Jianshu Li, Yulan Guo, Junliang Xing, et~al.
\newblock Driveworld: 4d pre-trained scene understanding via world models for
  autonomous driving.
\newblock In \emph{2024 IEEE/CVF Conference on Computer Vision and Pattern
  Recognition (CVPR)}, pages 15522--15533. IEEE, 2024.

\bibitem[Mur-Labadia et~al.(2026)Mur-Labadia, Muckley, Bar, Assran, Sinha,
  Rabbat, LeCun, and Ballas]{murlabadia2026vjepa21}
Lorenzo Mur-Labadia, Matthew Muckley, Amir Bar, Mido Assran, Koustuv Sinha,
  Mike Rabbat, Yann LeCun, and Nicolas Ballas.
\newblock V-jepa 2.1: Unlocking dense features in video self-supervised
  learning.
\newblock In \emph{European Conference on Computer Vision}, pages 671--689.
  Springer, 2026.

\bibitem[Song et~al.(2026)Song, Liu, Pan, Liao, Guo, Yang, Zhang, Xu, Jia, and
  Luo]{song2025breaking}
Ziying Song, Lin Liu, Hongyu Pan, Bencheng Liao, Mingzhe Guo, Lei Yang,
  Yongchang Zhang, Shaoqing Xu, Caiyan Jia, and Yadan Luo.
\newblock Diver: Reinforced diffusion breaks imitation bottlenecks in
  end-to-end autonomous driving.
\newblock \emph{IEEE Transactions on Pattern Analysis and Machine
  Intelligence}, 2026.

\bibitem[Su et~al.(2024)Su, Wu, Yang, and Guan]{su2024egofsd}
Haisheng Su, Wei Wu, Zhenjie Yang, and Isabel Guan.
\newblock Egofsd: Ego-centric fully sparse paradigm with uncertainty denoising
  and iterative refinement for efficient end-to-end self-driving.
\newblock \emph{arXiv preprint arXiv:2409.09777}, 2024.

\bibitem[Sun et~al.(2026)Sun, Lin, Chen, Pei, Li, Shi, and
  Zheng]{sun2026sparsedrivev2}
Wenchao Sun, Xuewu Lin, Keyu Chen, Zixiang Pei, Xiang Li, Yining Shi, and Sifa
  Zheng.
\newblock Sparsedrivev2: Scoring is all you need for end-to-end autonomous
  driving.
\newblock In \emph{European Conference on Computer Vision}, pages 446--463.
  Springer, 2026.

\bibitem[Tan et~al.(2026)Tan, Chitta, Chen, Tian, You, Wang, Luo, Cao,
  Kr{\"a}henb{\"u}hl, Pavone, et~al.]{tan2026latentcot}
Shuhan Tan, Kashyap Chitta, Yuxiao Chen, Ran Tian, Yurong You, Yan Wang, Wenjie
  Luo, Yulong Cao, Philipp Kr{\"a}henb{\"u}hl, Marco Pavone, et~al.
\newblock Latent chain-of-thought world modeling for end-to-end autonomous
  driving.
\newblock In \emph{Proceedings of the IEEE/CVF Conference on Computer Vision
  and Pattern Recognition}, pages 39724--39733, 2026.

\bibitem[Wang et~al.(2026{\natexlab{a}})Wang, Hua, Liu, Xing, Zhang, Ma, Chen,
  Ye, Chen, and Zhang]{wang2026beyonddrive}
Junli Wang, Zhihua Hua, Xueyi Liu, Zebin Xing, Wei Zhang, Kun Ma, Guang Chen,
  Hangjun Ye, Long Chen, and Qichao Zhang.
\newblock Beyond imitation: Learning safe end-to-end autonomous driving from
  hard negatives.
\newblock In \emph{European Conference on Computer Vision}, pages 224--241.
  Springer, 2026{\natexlab{a}}.

\bibitem[Wang et~al.(2026{\natexlab{b}})Wang, Zheng, Chen, Li, Zhang, Xing,
  Zhang, Li, Qian, Yang, et~al.]{wang2026latentwam}
Linbo Wang, Yupeng Zheng, Qiang Chen, Shiwei Li, Yichen Zhang, Zebin Xing,
  Qichao Zhang, Xiang Li, Deheng Qian, Pengxuan Yang, et~al.
\newblock {Latent-WAM: Latent World Action Modeling for End-to-End Autonomous
  Driving}.
\newblock \emph{arXiv preprint arXiv:2603.24581}, 2026{\natexlab{b}}.

\bibitem[Wang et~al.(2026{\natexlab{c}})Wang, Yang, Bai, Zhang, Liu, Zheng,
  Long, Lu, and Lu]{wang2026drivejepa}
Linhan Wang, Zichong Yang, Chen Bai, Guoxiang Zhang, Xiaotong Liu, Xiaoyin
  Zheng, Xiao-Xiao Long, Chang-Tien Lu, and Cheng Lu.
\newblock Drive-jepa: Video jepa meets multimodal trajectory distillation for
  end-to-end driving.
\newblock \emph{arXiv preprint arXiv:2601.22032}, 2026{\natexlab{c}}.

\bibitem[Wang et~al.(2024)Wang, He, Fan, Li, Chen, and Zhang]{wang2024driving}
Yuqi Wang, Jiawei He, Lue Fan, Hongxin Li, Yuntao Chen, and Zhaoxiang Zhang.
\newblock Driving into the future: Multiview visual forecasting and planning
  with world model for autonomous driving.
\newblock In \emph{2024 IEEE/CVF Conference on Computer Vision and Pattern
  Recognition (CVPR)}, pages 14749--14759. IEEE, 2024.

\bibitem[Wang et~al.(2026{\natexlab{d}})Wang, Li, Wang, Zhang, Li, Chen, Wang,
  and Zhang]{wang2025unified}
Yuqi Wang, Xinghang Li, Wenxuan Wang, Junbo Zhang, Yingyan Li, Yuntao Chen,
  Xinlong Wang, and Zhaoxiang Zhang.
\newblock Unified vision-language-action model.
\newblock In \emph{International Conference on Learning Representations},
  volume 2026, pages 80929--80944, 2026{\natexlab{d}}.

\bibitem[Wozniak et~al.(2026)Wozniak, Liu, Cai, and Jensfelt]{wozniak2025prix}
Maciej Wozniak, Lianhang Liu, Yixi Cai, and Patric Jensfelt.
\newblock Prix: Learning to plan from raw pixels for end-to-end autonomous
  driving.
\newblock \emph{IEEE Robotics and Automation Letters}, 11\penalty0
  (5):\penalty0 6400--6407, 2026.

\bibitem[Wu et~al.(2022)Wu, Jia, Chen, Yan, Li, and Qiao]{wu2022tcp}
Penghao Wu, Xiaosong Jia, Li~Chen, Junchi Yan, Hongyang Li, and Yu~Qiao.
\newblock Trajectory-guided control prediction for end-to-end autonomous
  driving: A simple yet strong baseline.
\newblock \emph{Advances in Neural Information Processing Systems},
  35:\penalty0 6119--6132, 2022.

\bibitem[Xiong et~al.(2026)Xiong, Guo, Li, Yan, Xu, Zhou, Chen, Sun, Wang, Ma,
  et~al.]{xiong2026recogdrive}
Kaixin Xiong, Xiangyu Guo, Fang Li, Sixu Yan, Gangwei Xu, Lijun Zhou, Long
  Chen, Haiyang Sun, Bing Wang, Kun Ma, et~al.
\newblock Recogdrive: A reinforced cognitive framework for end-to-end
  autonomous driving.
\newblock In \emph{International Conference on Learning Representations},
  volume 2026, pages 157518--157556, 2026.

\bibitem[Yang et~al.(2026)Yang, Chen, Huang, and Li]{yang2026autojepa}
Jiwei Yang, Zhengxian Chen, Chaosheng Huang, and Jun Li.
\newblock Auto-jepa: A latent world model of continuous intent for end-to-end
  autonomous driving.
\newblock \emph{arXiv preprint arXiv:2607.29031}, 2026.

\bibitem[Yao et~al.(2026)Yao, Li, Lan, Wang, Sun, Alvarez, and
  Wu]{yao2026drivesuprim}
Wenhao Yao, Zhenxin Li, Shiyi Lan, Zi~Wang, Xinglong Sun, Jose~M Alvarez, and
  Zuxuan Wu.
\newblock Drivesuprim: Towards precise trajectory selection for end-to-end
  planning.
\newblock In \emph{Proceedings of the AAAI Conference on Artificial
  Intelligence}, volume~40, pages 11910--11918, 2026.

\bibitem[Zhai et~al.(2023)Zhai, Feng, Du, Mao, Liu, Tan, Zhang, Ye, and
  Wang]{zhai2023admlp}
Jiang-Tian Zhai, Ze~Feng, Jinhao Du, Yongqiang Mao, Jiang-Jiang Liu, Zichang
  Tan, Yifu Zhang, Xiaoqing Ye, and Jingdong Wang.
\newblock Rethinking the open-loop evaluation of end-to-end autonomous driving
  in nuscenes.
\newblock \emph{arXiv preprint arXiv:2305.10430}, 2023.

\bibitem[Zhang et~al.(2025)Zhang, Song, Jin, and Zhang]{zhang2025bridging}
Bozhou Zhang, Nan Song, Xin Jin, and Li~Zhang.
\newblock Bridging past and future: End-to-end autonomous driving with
  historical prediction and planning.
\newblock In \emph{2025 IEEE/CVF Conference on Computer Vision and Pattern
  Recognition (CVPR)}, pages 6854--6863. IEEE, 2025.

\bibitem[Zhang et~al.(2026{\natexlab{a}})Zhang, Song, Zhu, Deng, Zhang,
  et~al.]{zhang2026future}
Bozhou Zhang, Nan Song, Xiatian Zhu, Jiankang Deng, Li~Zhang, et~al.
\newblock Future-aware end-to-end driving: Bidirectional modeling of trajectory
  planning and scene evolution.
\newblock \emph{Advances in Neural Information Processing Systems},
  38:\penalty0 10204--10229, 2026{\natexlab{a}}.

\bibitem[Zhang et~al.(2026{\natexlab{b}})Zhang, Li, and Li]{zhang2026idol}
Chenghao Zhang, Timin Li, and Dongmei Li.
\newblock Idol: Inverse-dynamics-guided future prediction for end-to-end
  autonomous driving.
\newblock \emph{arXiv preprint arXiv:2605.31476}, 2026{\natexlab{b}}.

\bibitem[Zheng et~al.(2024)Zheng, Song, Guo, Zhang, and Chen]{zheng2024genad}
Wenzhao Zheng, Ruiqi Song, Xianda Guo, Chenming Zhang, and Long Chen.
\newblock Genad: Generative end-to-end autonomous driving.
\newblock In \emph{European Conference on Computer Vision}, pages 87--104.
  Springer, 2024.

\bibitem[Zheng et~al.(2025)Zheng, Yang, Xing, Zhang, Zheng, Gao, Li, Zhang,
  Xia, Jia, et~al.]{zheng2025world4drive}
Yupeng Zheng, Pengxuan Yang, Zebin Xing, Qichao Zhang, Yuhang Zheng, Yinfeng
  Gao, Pengfei Li, Teng Zhang, Zhongpu Xia, Peng Jia, et~al.
\newblock World4drive: End-to-end autonomous driving via intention-aware
  physical latent world model.
\newblock In \emph{2025 IEEE/CVF International Conference on Computer Vision
  (ICCV)}, pages 28632--28642. IEEE, 2025.

\bibitem[Zou et~al.(2025)Zou, Chen, Liao, Zheng, Song, Zhang, Zhang, Liu, and
  Wang]{zou2025diffusiondrivev2}
Jialv Zou, Shaoyu Chen, Bencheng Liao, Zhiyu Zheng, Yuehao Song, Lefei Zhang,
  Qian Zhang, Wenyu Liu, and Xinggang Wang.
\newblock Diffusiondrivev2: Reinforcement learning-constrained truncated
  diffusion modeling in end-to-end autonomous driving.
\newblock \emph{arXiv preprint arXiv:2512.07745}, 2025.

\end{thebibliography}
\bibliographystyle{plainnat}

\clearpage
\appendix
\raggedbottom

\section*{Appendix}

Table of contents:
\begin{itemize}
    \item Appendix~\ref{app:representation_mechanism} analyzes the mechanisms
    of representation adaptation and dense future-feature prediction.
    \item Appendix~\ref{app:additional_analysis} collects additional ablations
    and efficiency measurements.
    \item Appendix~\ref{app:evaluation} defines the datasets and metrics.
    \item Appendix~\ref{app:limitations} discusses the scope and limitations of
    the current study.
    \item Appendix~\ref{app:implementation} provides notation, architecture,
    training, and inference details.
\end{itemize}

\section{Representation and Future-Prediction Analysis}
\label{app:representation_mechanism}

This section examines how Stage~1 LoRA adaptation reshapes the shared visual representation and how the dense JEPA decoder captures future scene evolution. Attention maps show spatial responses relative to uniform attention; PC1 maps show a shared projection of the high-dimensional features.

\begin{figure}[h]
    \centering
    \methodfigure
    {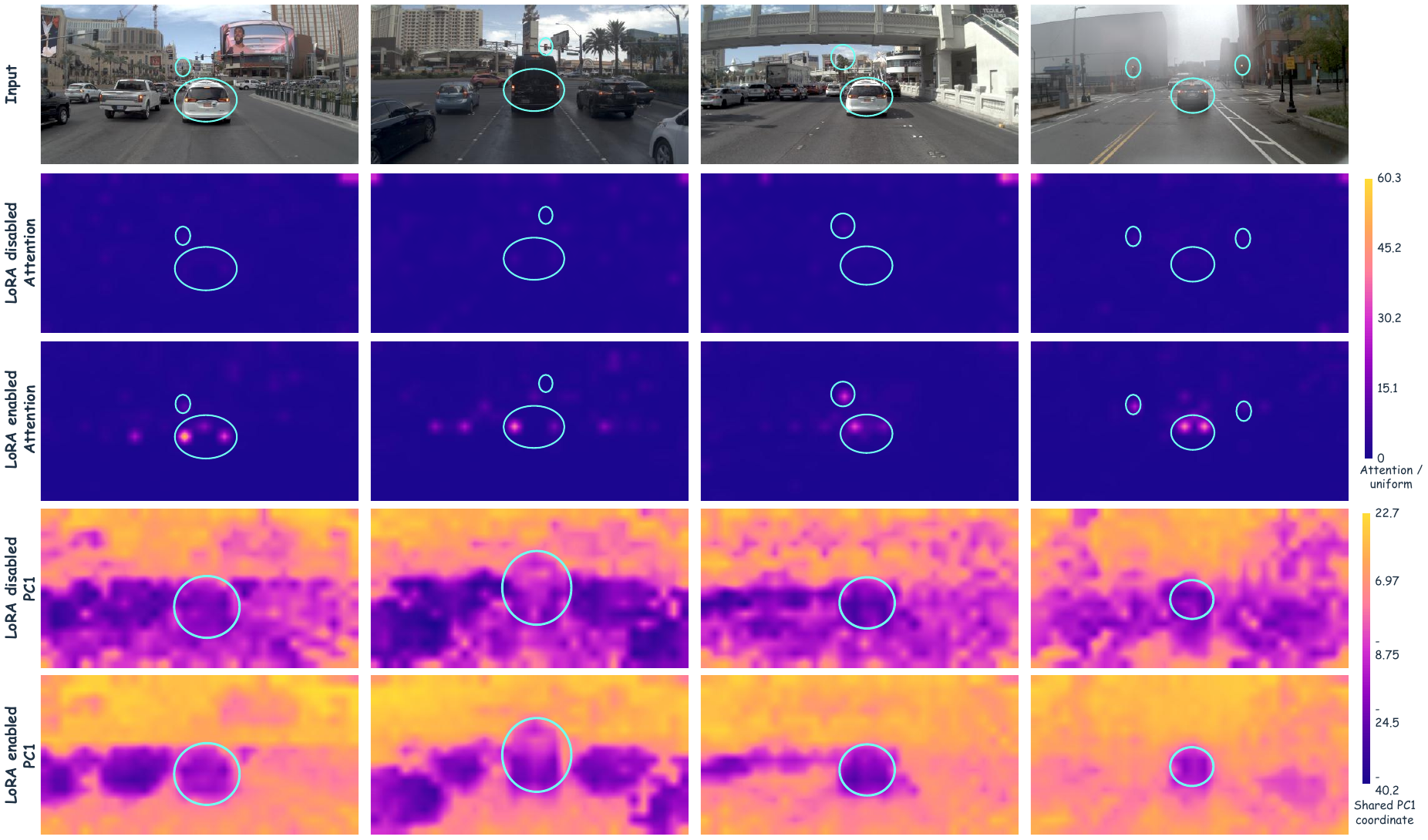}
    {0.90\textwidth}
    {8.0cm}
    \caption{\textbf{Spatial reorganization under LoRA adaptation.} Identical input frames are processed with the learned adapters disabled and enabled in the same trained model. Attention maps show a redistribution of spatial response, including clearer local responses near several annotated vehicles and signals; the color scale is relative to uniform attention. Shared PC1 projections show complementary changes in feature organization while preserving broader scene context.}
    \label{fig:lora_feature_comparison}
\end{figure}

\subsection{Spatial Reorganization Under LoRA Adaptation}

Fig.~\ref{fig:lora_feature_comparison} compares identical input frames with the learned LoRA adapters disabled and enabled in the same trained model. With adaptation enabled, localized attention responses become more visible around preceding vehicles and traffic signals. Several foreground regions become more distinct in the shared PC1 projections, while broader scene context remains visible. Together, these changes characterize task-dependent adaptation under the joint predictive and planning objectives.

\subsection{Dense Future-Feature Evolution}

\begin{figure}
    \centering
    \methodfigure
    {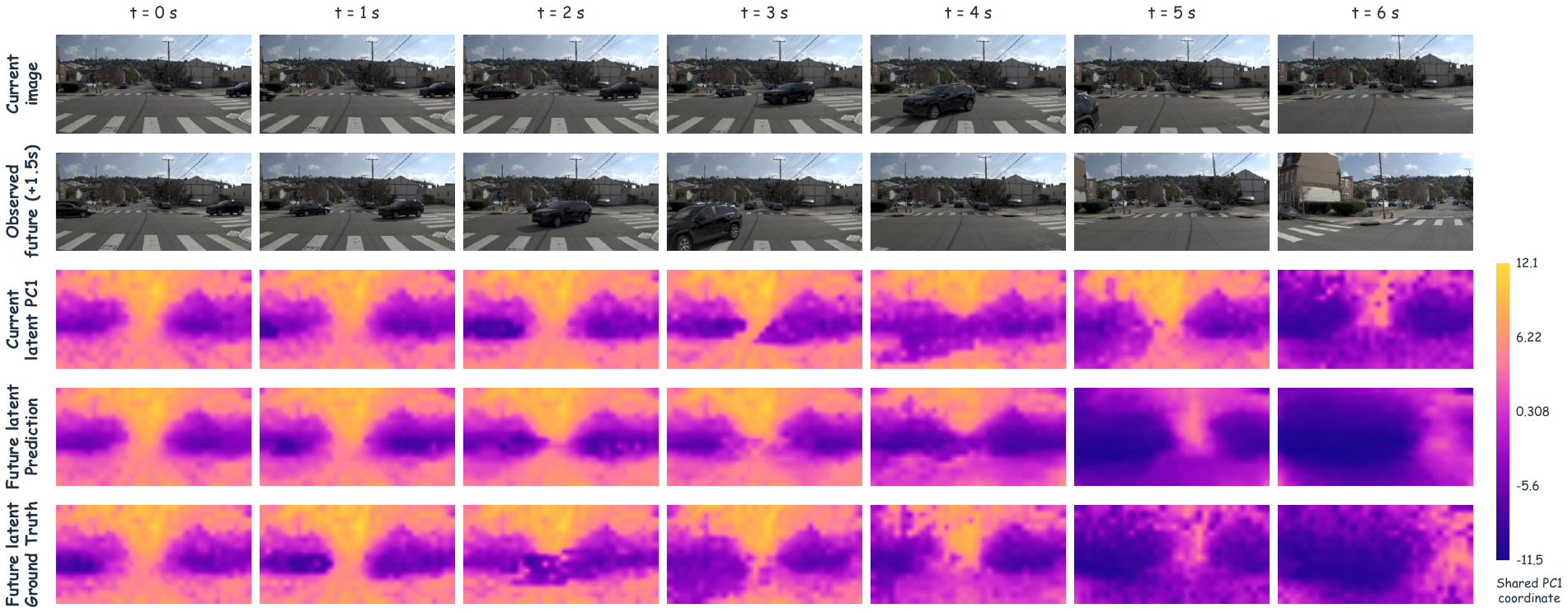}
    {0.95\textwidth}
    {6.0cm}
    \caption{\textbf{Future-feature prediction during a crossing interaction.} Each column is an independent 1.5-second-ahead prediction. During the transition, some predicted structures remain closer to the current map than to the target, whereas the $t=5$ prediction captures part of the target's broader shift toward lower PC1 values. The dense branch therefore follows some coarse future changes but incompletely recovers local and transitional structure.}
    \label{fig:crossing_future_latent}
\end{figure}

\begin{figure}[tbp]
    \centering
    \methodfigure
    {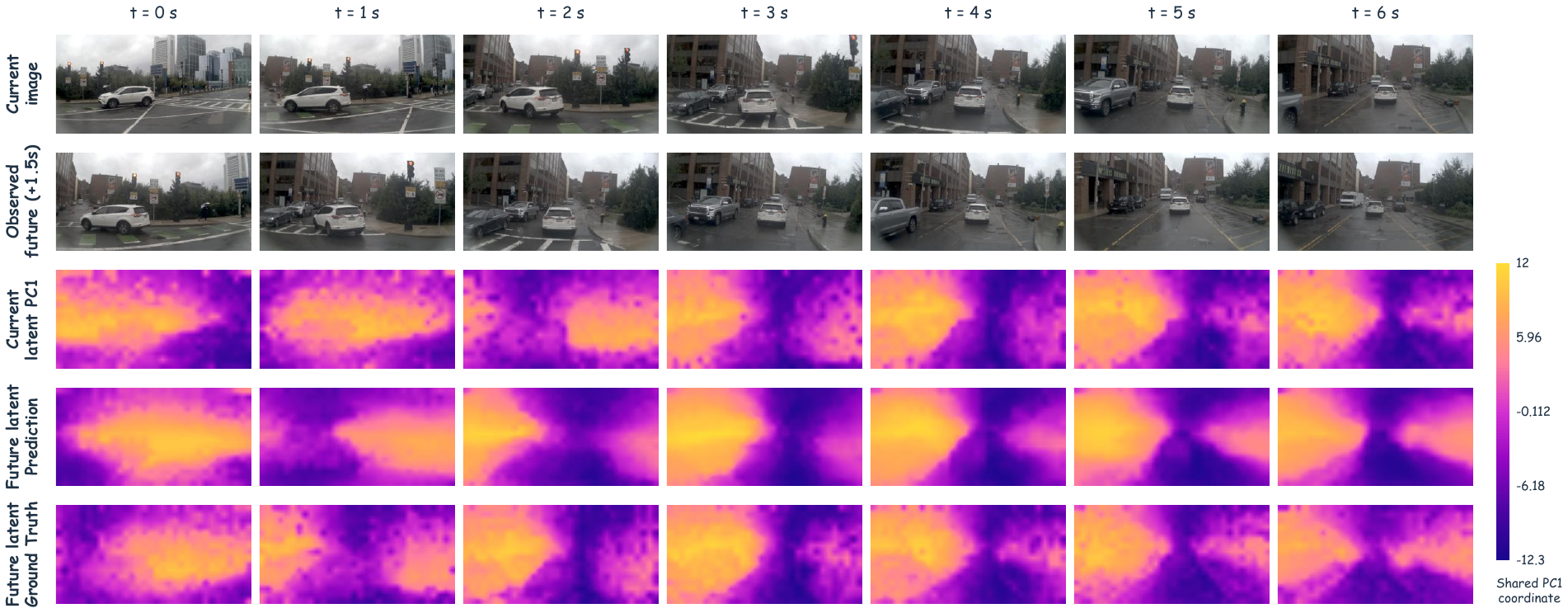}
    {0.95\textwidth}
    {6.0cm}
    \caption{\textbf{Future-feature reorganization during a turn.} Each column is an independent 1.5-second-ahead prediction. At $t=2$, a dominant higher-valued region shifts from the right in the current map to the left in both prediction and target.}
    \label{fig:suv_future_latent}
\end{figure}

Figs.~\ref{fig:crossing_future_latent} and~\ref{fig:suv_future_latent} provide two examples from the separately trained 1.5-second visualization checkpoint; quantitative evaluation uses the 0.5-second checkpoint. Each column is an independent prediction. From top to bottom, the rows show the current image, observed future image, current latent, dense decoder output, and EMA target future tokens. A shared PC1 projection enables spatial comparison across representations. The crossing example emphasizes local interaction changes, and the turning example emphasizes viewpoint-dependent spatial reorganization.

\paragraph{Crossing interaction.}
In Fig.~\ref{fig:crossing_future_latent}, the current, predicted, and target maps share substantial global structure in the earlier columns. At $t=2$ and $t=4$ seconds, parts of the prediction remain closer to the current map. At $t=5$ seconds, the prediction follows the target's shift toward a predominantly lower-valued PC1 layout. The dense branch captures coarse future evolution, with residual errors in local and transitional structure.

\paragraph{Turning sequence.}
Fig.~\ref{fig:suv_future_latent} shows spatial reorganization during a turn. At $t=2$ seconds, the dominant higher-valued region shifts from the right in the current map to the left in both prediction and target. Later columns show higher-valued side regions around a lower-valued central region. The prediction captures these broad changes; the emerging left-side structure at $t=1$ and finer target variation remain less accurately recovered.

\section{Additional Ablations and Efficiency}
\label{app:additional_analysis}

This section groups supporting analyses that complement the matched
ablations in Section~\ref{sec:ablations}.

\subsection{Visual Backbone Ablations}
\label{app:backbone_scale}

We compare V-JEPA~2.1 backbones in the preliminary planning baseline that preceded the compact candidate-conditioned architecture. Table~\ref{tab:early_vjepa_scale} reports individual runs with end-to-end fine-tuning (E2E) and frozen encoders on NAVSIM-v1 Full Navtest (12,146 valid scenes).

\begin{table}[H]
\centering
\caption{\textbf{Backbone comparison in the preliminary planning baseline.} E2E and Frozen denote end-to-end adaptation and a fixed encoder. ViT-g and ViT-G are distinct checkpoint variants. PDMS is reported on a 0--100 scale.}
\label{tab:early_vjepa_scale}
\begin{tabular}{lrr}
\toprule
Backbone & E2E & Frozen \\
\midrule
ViT-B (\texttt{vitb}) & 86.9958 & 85.1258 \\
ViT-L (\texttt{vitl}) & \textbf{88.1473} & \textbf{85.5403} \\
ViT-g (\texttt{vitg}) & 87.1119 & 84.6653 \\
ViT-G (\texttt{vitG}) & 87.9113 & 81.4653 \\
\bottomrule
\end{tabular}
\end{table}

All runs use 40 training epochs. The global batch size is 64, except for frozen ViT-G, which uses 128. ViT-B/L are distilled checkpoints, while ViT-g/G follow their respective pretraining recipes. ViT-L achieves the highest PDMS in both adaptation settings, and end-to-end adaptation outperforms freezing for every tested variant.

\subsection{Inference Latency and Candidate-Count Scaling}
\label{app:forward_time}

Table~\ref{tab:forward_time_comparison} compares forward-pass time with DiffusionDrive and DrivoR. For \method{}, increasing the candidate count from 1 to 64 keeps time within 36.52--39.00~ms, and 128 candidates require 42.57~ms, indicating effective parallel processing. At the default $N=32$, \method{} takes 37.24~ms: 12.56~ms slower than DiffusionDrive but 258.32~ms faster than DrivoR.

\begin{table}[H]
    \centering
    \small
    \caption{\textbf{Inference latency and candidate-count scaling.} Reported forward-pass time in milliseconds; lower is better.}
    \label{tab:forward_time_comparison}
    \setlength{\tabcolsep}{8.0pt}
    \begin{tabular}{lcc}
        \toprule
        Method & Candidates & Forward time (ms) $\downarrow$ \\
        \midrule
        DiffusionDrive~\cite{liao2025diffusiondrive} & -- & 24.68 \\
        DrivoR~\cite{kirby2026drivor} & -- & 295.56 \\
        \midrule
        & 1   & 38.11 \\
        & 4   & 36.89 \\
        & 8   & 36.52 \\
        & 16  & 39.00 \\
        \highlightrow
        \method{} & 32  & 37.24 \\
        & 64  & 36.97 \\
        & 128 & 42.57 \\
        \bottomrule
    \end{tabular}
\end{table}

\FloatBarrier
\section{Evaluation Protocol and Metrics}
\label{app:evaluation}

We evaluate on the NAVSIM-v1 and NAVSIM-v2 \texttt{navtest} splits. NAVSIM-v1 contains 12,146 driving scenarios~\cite{dauner2024navsim}. Unless stated otherwise, all metric values are computed on the native $[0,1]$ scale and reported after multiplication by 100. The \method{} result in Table~\ref{tab:benchmark_navsim_v2} uses the corrected human-filter aggregation, which recomputes the multiplicative and weighted terms after filtering.

NAVSIM-v1 reports No-at-Fault Collision (NC), Drivable Area Compliance (DAC), Ego Progress (EP), Time to Collision (TTC), and Comfort (C). NC and DAC measure collision avoidance and road compliance; TTC indicates whether the trajectory remains free of near-term collision risk; and C checks motion smoothness. EP is derived from the nonnegative start-to-end progress of the simulated ego center projected onto the route centerline and is normalized within the official proposal-scoring procedure using safety-masked progress. It is therefore not the Euclidean or rendered length of the trajectory segment shown in a bird's-eye-view panel; a visually short plotted curve can still receive EP~$=1$. Safety failures affect the aggregate separately through the multiplicative NC and DAC terms. The per-scenario score is
\begin{equation}
\mathrm{PDMS}
=
\mathrm{NC}\,\mathrm{DAC}
\frac{5\,\mathrm{EP}+5\,\mathrm{TTC}+2\,\mathrm{C}}{12}.
\label{eq:pdms_composition}
\end{equation}
DDC is reported as an auxiliary diagnostic in the Stage~2 checkpoint analysis; NAVSIM-v1 PDMS uses the factors in Eq.~\ref{eq:pdms_composition}.

NAVSIM-v2 extends the evaluation with DDC, Traffic Light Compliance (TLC), Lane Keeping (LK), History Comfort (HC), and Extended Comfort (EC). These respectively measure correct driving direction, red-light compliance, lane adherence, continuity with the observed motion history, and consistency across consecutive predictions. Let the tildes denote the official human-reference filtering applied by NAVSIM-v2. Its per-scenario score is
\begin{equation}
\mathrm{EPDMS}
=
\widetilde{\mathrm{NC}}\,
\widetilde{\mathrm{DAC}}\,
\widetilde{\mathrm{DDC}}\,
\widetilde{\mathrm{TLC}}
\frac{
5\widetilde{\mathrm{EP}}+5\widetilde{\mathrm{TTC}}
+2\widetilde{\mathrm{LK}}+2\widetilde{\mathrm{HC}}+2\widetilde{\mathrm{EC}}
}{16}.
\label{eq:epdms_composition}
\end{equation}

\section{Limitations}
\label{app:limitations}

Dense future-feature supervision is restricted to the expert-matched candidate. Alternative candidates receive planning-derived labels, and their visual outcomes remain unobserved. Expert matching uses minimum ADE without a rejection threshold; the matching-error distribution and its effect on prediction accuracy remain unmeasured. The dense-feature visualizations assess the complete decoding path, leaving the future information retained in the compact tokens to be quantified separately.

Evaluation covers NAVSIM and Bench2Drive. Broader geographic, sensor, weather, and rare-event coverage is needed to assess deployment robustness. The preliminary backbone comparison uses individual runs with different checkpoint recipes and one batch-size difference; controlled multi-seed comparisons remain future work.

Safety-focused post-training improves safety and compliance with a modest reduction in progress. NAVSIM-v2 additionally uses training-free temporal-consistency reranking.

\section{Model and Training Details}
\label{app:implementation}

This appendix first consolidates the notation and then describes the
architecture, training configuration, inference procedure, and layer-wise
implementation of the final model.

\subsection{Notation}
\label{app:notation}

Table~\ref{tab:notation} summarizes the symbols shared by the main paper and
the detailed formulation below.

\begin{table}[tbp]
    \centering
    \caption{\textbf{Notation.} Symbols used in the main formulation and implementation details. Compact candidate-conditioned tokens $\widehat z_i$ are distinguished from dense future-feature predictions $\widetilde Z_{i^{\mathrm{exp}}}$.}
    \label{tab:notation}
    \begin{tabular}{@{}p{0.28\columnwidth}p{0.64\columnwidth}@{}}
        \toprule
        Symbol & Meaning \\
        \midrule
        $X_t$, $X_{t+\Delta}$
        & Current and observed future visual inputs. \\
        $u_t$
        & Ego-state input used by the trajectory generator. \\
        $\mathcal{T}=\{\tau_i\}_{i=1}^{N}$
        & Set of $N$ candidate ego trajectories. \\
        $\tau^{\mathrm{exp}}$, $\tau_j^{-}$, $\tau^\star$
        & Expert, hard-negative, and finally selected trajectories. \\
        $Q_i$
        & Native pose-token sequence produced with candidate $\tau_i$. \\
        $E_\theta$, $E_{\bar\theta}$
        & Online encoder and EMA target encoder. \\
        $Z_t$, $Z_{t+\Delta}$
        & Current scene tokens and observed future target tokens. \\
        $a_i=E_\tau(\tau_i)$
        & Candidate query tokens formed from $Q_i$ and the geometry of $\tau_i$. \\
        $\widetilde Z_{i^{\mathrm{exp}}}=D_\omega(Z_t,\widehat z_{i^{\mathrm{exp}}})$
        & Dense token prediction decoded from the expert-matched compact tokens for Stage~1 feature alignment. \\
        $\widehat z_i=P_\phi(a_i,Z_t)$
        & Compact candidate-future token sequence used by the scorer. \\
        $S_\psi$
        & Shared trajectory scorer conditioned on $Z_t$, $a_i$, and $\widehat z_i$. \\
        $i^{\mathrm{exp}}$
        & Index minimizing planar $(x,y)$ $\ell_2$ ADE to the expert trajectory; yaw is excluded. \\
        $\mathbf q_i$, $\widehat{\mathbf q}_i$
        & Target and predicted planning-factor vectors. \\
        $s_i$, $\widehat s_i$
        & Target and predicted overall trajectory utilities. \\
        $N$, $T$, $M$, $D$
        & Numbers of candidates, trajectory steps, scene tokens, and token dimensions. \\
        $\mu$
        & Momentum coefficient for the EMA target encoder. \\
        $\mathcal L_{\mathrm{pred}}$, $\mathcal L_{\mathrm{factor}}$
        & Future-prediction and planning-factor losses. \\
        $\mathcal L_{\mathrm{score}}$, $\mathcal L_{\mathrm{rank}}$
        & Utility-regression and pairwise-ranking losses. \\
        $\mathcal L_1$, $\mathcal L_2$
        & Stage~1 joint-adaptation and Stage~2 safety-refinement objectives. \\
        $\lambda_{\cdot}$
        & Weights used to combine the training losses. \\
        \bottomrule
    \end{tabular}
\end{table}

\subsection{Model Architecture}
\label{app:architecture_impl}

We use a pretrained V-JEPA~2.1 visual encoder with rank-32 LoRA
adapters applied to the query and value projections of each
Transformer block. The original backbone parameters remain frozen
during joint training.

The trajectory generator produces $N=32$ candidate trajectories, matching
the candidate-set notation $\mathcal T=\{\tau_i\}_{i=1}^{N}$ used in the
main text.
For each candidate, a trajectory-conditioned predictor constructs
a compact future representation. The shared scorer combines this
representation with the candidate trajectory and current scene
features to predict a scalar planning score.

During joint training, an exponential moving average (EMA) teacher
provides future visual targets from the two-frame window
$[t,t+0.5\,\mathrm{s}]$. A full-latent decoder reconstructs the
teacher representation from the ground-truth-matched candidate.
The reconstruction objective uses an L1 loss with weight $0.1$.
The EMA teacher and full-latent reconstruction decoder are used
only during training; candidate-conditioned future prediction
remains active at inference.

\begin{table}[tbp]
\centering
\caption{\textbf{Default architecture and Stage~1 training configuration.} Unless stated otherwise, the model uses 32 generated candidates, rank-32 LoRA, and a 0.5-second predictive target. External candidates are used only for training supervision.}
\label{tab:architecture_training}
\begin{tabular}{lc}
\toprule
Parameter & Value \\
\midrule
Visual encoder & V-JEPA~2.1 \\
LoRA rank & 32 \\
LoRA target projections & Query and value \\
Generated candidates & 32 \\
External counterfactual candidates & 128 \\
Future prediction horizon & $0.5\,\mathrm{s}$ \\
Future reconstruction weight & 0.1 \\
Joint-training epochs & 30 \\
Initial learning rate & $10^{-4}$ \\
Number of GPUs & 8 \\
Batch size per GPU & 8 \\
Global batch size & 64 \\
Mixed precision & bf16 \\
\bottomrule
\end{tabular}
\end{table}

Table~\ref{tab:architecture_training} describes the quantitative checkpoint. The 1.5-second feature visualizations use a separately trained checkpoint with the same architecture and training setup except for the target horizon. Each evaluated checkpoint uses a fixed horizon. Stage~1 trains the proposal generator, LoRA adapters, compact predictor, dense decoder, and scorer while keeping the pretrained backbone weights fixed. Safety refinement starts from the Stage~1 checkpoint at epoch 15.

\subsection{Joint Training}
\label{app:joint_training}

Using the notation of Eq.~\ref{eq:total}, the implementation-level
decomposition of the Stage~1 objective is
\begin{equation}
\begin{aligned}
\mathcal{L}_{1}^{\mathrm{impl}}
={}&
\mathcal{L}_{\mathrm{traj}}
+0.1\,\mathcal{L}_{\mathrm{pred}}
+\mathcal{L}_{\mathrm{factor}}^{\mathrm{gen}}
+\mathcal{L}_{\mathrm{score}}^{\mathrm{gen}}
\\
&+\lambda_{\mathrm{ext}}(e)\left(
\mathcal{L}_{\mathrm{factor}}^{\mathrm{ext}}
+\mathcal{L}_{\mathrm{score}}^{\mathrm{ext}}
\right)
+\mathcal{L}_{\mathrm{aux}},
\end{aligned}
\end{equation}
where $\mathcal{L}_{\mathrm{pred}}$ is the dense future-feature loss in
Eq.~\ref{eq:pred}. For both generated and external candidates,
$\mathcal{L}_{\mathrm{factor}}$ applies binary cross-entropy to the five
planning-factor outputs and $\mathcal{L}_{\mathrm{score}}$ applies binary
cross-entropy to the learned utility output. Both have unit weight within
each candidate group. Pairwise ranking is disabled in Stage~1.
$\mathcal{L}_{\mathrm{traj}}$ denotes
the inherited proposal loss, while $\mathcal{L}_{\mathrm{aux}}$ contains
the inherited agent and area auxiliary objectives.

Each scene supplies 128 external counterfactual trajectories,
comprising 64 local-hard candidates and 64 balanced candidates.
Their coordinates and current-scene features are detached before
entering the external scoring branch. Consequently, the external
loss updates the candidate-future modules and shared scorer
without updating the trajectory generator or visual backbone
through this branch.

Using zero-based epoch indices, the counterfactual weight is
\begin{equation}
\lambda_{\mathrm{ext}}(e)=
\begin{cases}
0, & 0 \leq e \leq 4,\\
0.05(e-4), & 5 \leq e \leq 9,\\
0.25, & 10 \leq e \leq 29.
\end{cases}
\label{eq:external_weight}
\end{equation}
Joint training runs for 30 epochs with an initial learning rate
of $10^{-4}$, a global batch size of 64, and bf16 mixed precision.

\subsection{Safety-Focused Post-Training}
\label{app:safety_posttraining}

We initialize safety-focused post-training from the joint-training
checkpoint at epoch 15. We freeze the visual encoder, including
its LoRA adapters, the trajectory generator, the full-latent
decoder, and the EMA teacher. Only the candidate-conditioned
future branch and the shared scorer are optimized.
The future reconstruction forward pass is retained, but its loss
weight is set to zero during this stage.

Each scene contains 32 generated trajectories and 128 external
counterfactual trajectories. The external set consists of
32 local-hard, 32 high-safe, 16 medium-score, 16 collision-failure,
8 drivable-area-failure, and 24 TTC-only-failure candidates.

A candidate is considered safe when
\begin{equation}
\mathrm{NC}\geq0.95,\qquad
\mathrm{DAC}\geq0.95,\qquad
\mathrm{TTC}\geq0.95.
\end{equation}
We supervise both the five planning-factor outputs and the learned utility.
Both use binary cross-entropy, and the utility-loss contribution of unsafe
candidates is multiplied by $4$.

For each candidate set, let $i^{+}$ denote the safe candidate
with the highest target final score, and let $\mathcal{U}_{8}$
contain unsafe candidates among the eight highest predicted
scores. The safety-ranking loss is
\begin{equation}
\mathcal{L}_{\mathrm{rank}}
=
\frac{1}{|\mathcal{U}_{8}|}
\sum_{j\in\mathcal{U}_{8}}
\max\!\left(0,\,
0.05-(\widehat s_{i^{+}}-\widehat s_j)\right),
\end{equation}
where $\widehat s_i$ is the predicted utility from
Eq.~\ref{eq:trajectory_selection}.
The loss is zero when no valid safe--unsafe pair exists.

The Stage~2 objective is
\begin{equation}
\begin{aligned}
\mathcal{L}_{2}
={}&
\mathcal{L}_{\mathrm{factor}}^{\mathrm{gen}}
+
\mathcal{L}_{\mathrm{score}}^{\mathrm{gen,w}}
+0.5\left(
\mathcal{L}_{\mathrm{factor}}^{\mathrm{ext}}
+\mathcal{L}_{\mathrm{score}}^{\mathrm{ext,w}}
\right)
\\
&+0.1\left(
\mathcal{L}_{\mathrm{rank}}^{\mathrm{gen}}
+0.5\,\mathcal{L}_{\mathrm{rank}}^{\mathrm{ext}}
\right).
\end{aligned}
\end{equation}

The following table maps the main-text objectives to their
implementation-level weights and gradient targets. Here ``generated''
denotes the 32 proposal-generator candidates and ``external'' denotes the
sampled counterfactual candidates.

\begin{table}[t]
\centering
\scriptsize
\setlength{\tabcolsep}{3.5pt}
\caption*{\textbf{Training-loss mapping across the two stages.} The Stage~1 external
weight $\lambda_{\mathrm{ext}}(e)$ follows the warm-up schedule in
Eq.~\ref{eq:external_weight}.}
\begin{tabularx}{\textwidth}{>{\raggedright\arraybackslash}p{2.6cm}>{\raggedright\arraybackslash}p{2.5cm}>{\raggedright\arraybackslash}p{2.5cm}>{\raggedright\arraybackslash}X >{\raggedright\arraybackslash}X}
\toprule
Loss & Stage~1 weight & Stage~2 weight & Supervision target & Updated modules \\
\midrule
Future-feature regression
& $0.1$
& $0$
& Expert-matched generated candidate
& Stage~1 dense decoder, compact predictor, encoder adapters, and proposal path through native pose tokens \\
Factor prediction
& $1.0$ generated; $\lambda_{\mathrm{ext}}(e)$ external
& $1.0$ generated; $0.5$ external
& Five planning factors for all generated and sampled external candidates
& Generated Stage~1 path: compact branch, scorer, encoder adapters, and proposal generator through $Q_i$; external Stage~1 and all Stage~2 paths: compact branch and scorer \\
Utility prediction
& $1.0$ generated; $\lambda_{\mathrm{ext}}(e)$ external
& $1.0$ generated; $0.5$ external; unsafe BCE $\times4$
& Learned-utility target for all generated and sampled external candidates
& Generated Stage~1 path: compact branch, scorer, encoder adapters, and proposal generator through $Q_i$; external Stage~1 and all Stage~2 paths: compact branch and scorer \\
Safe--unsafe ranking
& $0$
& $0.1$ generated; $0.05$ external
& Highest-target safe candidate versus unsafe candidates among predicted Top-$8$
& Stage~2 compact branch and scorer \\
Inherited proposal loss
& $1.0$
& $0$
& Generated trajectories against the expert trajectory
& Proposal generator and its trainable upstream path \\
Inherited auxiliary losses
& $1.0$ outer weight
& $0$
& Agent and area targets from the base planner
& Corresponding base-planner modules \\
\bottomrule
\end{tabularx}
\end{table}

Post-training runs for three epochs with a learning rate of
$10^{-6}$ and the same batch size and precision as joint training.
The final released model uses the checkpoint after two completed
post-training epochs, corresponding to zero-based epoch 1.

\begin{table}[tbp]
\centering
\caption{\textbf{Stage~2 safety-refinement configuration.} The visual encoder, proposal generator, dense decoder, and EMA target are fixed; only the compact predictor and scorer are optimized. Three epochs are run, the model after two completed epochs is retained, and the future-feature loss has zero weight.}
\label{tab:posttraining_hparams}
\begin{tabular}{lc}
\toprule
Parameter & Value \\
\midrule
Learning rate & $10^{-6}$ \\
Post-training epochs & 3 \\
Retained checkpoint & Epoch 1 (zero-based) \\
Global batch size & 64 \\
Generated factor/utility loss weight & 1.0 \\
External factor/utility loss weight & 0.5 \\
Unsafe utility-BCE multiplier & 4.0 \\
Safety-ranking weight & 0.1 \\
Safety-ranking Top-$K$ & 8 \\
Safety-ranking margin & 0.05 \\
Future reconstruction loss weight & 0 \\
\bottomrule
\end{tabular}
\end{table}

\subsection{Inference and Temporal Reranking}
\label{app:inference_impl}

For NAVSIM v1, inference selects the candidate with the highest
learned final score. For the corrected NAVSIM-v2 evaluation, we
add training-free temporal-consistency reranking:
\begin{equation}
\widetilde{s}_{t,i}
=
0.875\,\widehat s_{t,i}
+
0.125\,\mathrm{EC}_{\mathrm{pair}}
\left(
\tau^\star_{t-1},
\tau_{t,i}
\right),
\end{equation}
where $\widehat s_{t,i}$ is the learned utility of the current candidate
and $\tau^\star_{t-1}$ is the trajectory actually
selected at the previous frame. The pairwise consistency term
is computed after aligning the trajectories across frames.
The selected trajectory is
\begin{equation}
\tau^\star_t
=
\tau_{t,i^\star},
\qquad
i^\star=\arg\max_{i\in\{1,\ldots,32\}}
\widetilde{s}_{t,i}.
\end{equation}

Temporal reranking requires no additional model training. The corrected
human filtering, metric computation, and aggregation protocol remain
unchanged by reranking.

\subsection{Layer-Wise Network Formulation}
\label{app:compact_network}

We formulate the compact candidate-conditioned model used to initialize the
safety-refined checkpoint, following the symbols in
Section~\ref{sec:trajectory_scoring}. Batch dimensions are omitted for
clarity. Let $X_t$ denote the available visual observations and $u_t$ the
ego state. We write the projected, flattened visual features as the scene
tokens
\begin{equation}
Z_t=\operatorname{Flatten}\!\left(\Pi(E_\theta(X_t))\right)
\in\mathbb{R}^{M\times D},
\qquad D=256,\quad M=HW,\quad(H,W)=(16,32).
\end{equation}
The projection $\Pi$ is absorbed into the abstract encoder notation $Z_t=E_\theta(X_t)$ in the main text. The spatial grid represents image features. The trajectory generator produces $N=32$ candidates and their native pose tokens:
\begin{equation}
\left(\{\tau_i\}_{i=1}^{N},
\{Q_i\}_{i=1}^{N}\right)
=G_\gamma(E_\theta(X_t),u_t),
\end{equation}
where $\tau_i\in\mathbb{R}^{T\times3}$ contains
$(x,y,\mathrm{yaw})$ coordinates, $Q_i\in\mathbb{R}^{T\times D}$,
and $T=8$. Visual features are computed once and shared across candidates.

\subsubsection{Trajectory-Conditioned Future Tokens}
\label{app:candidate_future_formulation}

Let $\operatorname{sg}$ denote stop-gradient and let $[\cdot;\cdot]$
denote channel-wise concatenation. The implementation of the abstract
trajectory embedding $a_i=E_\tau(\tau_i)$ and the scene memory are
\begin{align}
a_i
&=\psi_f\!\left([\operatorname{LN}(Q_i);
\eta_f(\operatorname{sg}(\tau_i))]\right)
+E_{\mathrm{pose}},\\
S_t
&=\operatorname{LN}\!\left(
Z_t+E_{\mathrm{spatial}}
\right),
\end{align}
where $\eta_f$ is a trajectory MLP and $\psi_f$ is a linear projection
followed by LayerNorm and GELU. The future predictor is
\begin{equation}
\widehat z_i
=P_\phi(a_i,Z_t)
=\operatorname{LN}\!\left(
D_f(a_i,S_t)\right)
\in\mathbb{R}^{T\times D}.
\end{equation}
Here $D_f$ comprises two pre-normalized Transformer decoder layers
with eight attention heads, feed-forward width 1024, and zero dropout.
Each layer applies candidate-token self-attention followed by
cross-attention to shared scene memory and a feed-forward update.
The candidate dimension is folded into the batch dimension during execution. The eight tokens are indexed by trajectory poses.

\subsubsection{Future Reconstruction and Gradient Boundaries}
\label{app:future_reconstruction_formulation}

During joint training, future reconstruction supervises the compact tokens of the candidate closest to the expert trajectory:
\begin{equation}
i^{\mathrm{exp}}=\arg\min_i\operatorname{ADE}(\tau_i,\tau^{\mathrm{exp}})
=\arg\min_i\frac{1}{T}\sum_{k=1}^{T}
 \left\|p_{i,k}-p^{\mathrm{exp}}_k\right\|_2,
\qquad p_{i,k}=(x_{i,k},y_{i,k}).
\end{equation}
Matching uses planar positions $(x,y)$. The selected index is detached, and the observed future target is assigned to the minimum-ADE candidate without a rejection threshold. A two-layer full-latent decoder maps its compact tokens to the spatial feature grid:
\begin{equation}
\widetilde Z_{i^{\mathrm{exp}}}
=D_\omega(Z_t,\widehat z_{i^{\mathrm{exp}}})
=\operatorname{LN}\!\left[
D_{\mathrm{full}}(S_t+E_{\mathrm{full}},
\widehat z_{i^{\mathrm{exp}}})\right]
\in\mathbb{R}^{M\times D}.
\end{equation}
No stop-gradient is applied to $\widehat z_{i^{\mathrm{exp}}}$ on this
path. Consequently, $\mathcal{L}_{\mathrm{pred}}$ updates the dense
decoder, the compact predictor, and the trainable encoder adapters during
Stage~1. Because the native pose tokens $Q_i$ are not detached, this loss
also has a gradient path into the proposal generator; only the generated
trajectory coordinates used by $\eta_f$ are detached. Compact tokens for
candidates other than $i^{\mathrm{exp}}$ receive no direct gradient from
$\mathcal{L}_{\mathrm{pred}}$.
For the quantitative setting $\Delta=0.5$ seconds, $X_{t+\Delta}$
denotes the observed target window $[I_t,I_{t+0.5}]$. The EMA encoder
produces the target tokens
\begin{equation}
Z_{t+\Delta}=\operatorname{sg}\!\left(
\operatorname{Flatten}\!\left(\Pi(E_{\bar\theta}(X_{t+\Delta}))\right)
\right).
\end{equation}
The implementation instantiates Eq.~\ref{eq:pred} with per-location
channel normalization:
\begin{equation}
\mathcal{L}_{\mathrm{pred}}
=\frac{1}{MD}\left\|
\operatorname{LN}_c(\widetilde Z_{i^{\mathrm{exp}}})
-\operatorname{LN}_c(Z_{t+\Delta})
\right\|_1,
\end{equation}
where $\operatorname{LN}_c$ normalizes channels independently at each
spatial location. The EMA teacher is updated with decay $\mu=0.99925$:
\begin{equation}
\bar{\theta}\leftarrow\mu\bar{\theta}+(1-\mu)\theta.
\end{equation}
Its parameter set includes the corresponding backbone projection.

For external candidates, both coordinates and scene features are detached before the candidate-future and scoring path. Safety-focused post-training freezes the backbone and trajectory generator.

\subsubsection{Candidate Scoring}
\label{app:scorer_formulation}

The scorer has its own trajectory embedding $\eta_s$ and fusion mapping
$\psi_s$. It combines native pose tokens, trajectory geometry, and
predicted future tokens:
\begin{align}
U_i
&=\psi_s\!\left([
\operatorname{LN}(Q_i);
\eta_s(\operatorname{sg}(\tau_i));
\operatorname{LN}(\widehat z_i)]\right),\\
h_i
&=\operatorname{MaxPool}_{k=1}^{T}
\left(\operatorname{TransformerEnc}(U_i)\right).
\end{align}
The temporal encoder contains one layer. With
$V_t=Z_t$, the final feature and multi-head outputs are
\begin{align}
r_i
&=\operatorname{LN}\!\left(
h_i+\operatorname{MHA}(
h_i,V_t,V_t)\right),\\
[\widehat{\mathbf q}_i;\widehat s_i]
&=\sigma\!\left(g_\psi(r_i)\right)\in\mathbb{R}^{6}.
\end{align}
The first five outputs are planning factors $\widehat{\mathbf q}_i$, and the sixth is the learned utility $\widehat s_i$ used for ranking. These parallel heads instantiate $S_\psi$ in Eq.~\ref{eq:trajectory_selection}. The scorer consumes compact tokens $\widehat z_i$ directly. The full-latent decoder and EMA teacher are omitted from inference.

\end{document}